\documentclass[10pt,letterpaper]{article}

\usepackage{times}
\usepackage[letterpaper,textwidth=5.5in,textheight=9in,centering]{geometry}
\usepackage[round,authoryear]{natbib}
\setcitestyle{authoryear,round,citesep={;},aysep={,},yysep={;}}
\usepackage{url}
\usepackage{amsmath,amssymb,amsfonts}
\usepackage{booktabs}
\usepackage{longtable}
\usepackage{graphicx}
\usepackage{enumitem}
\usepackage{microtype}
\usepackage{placeins}
\usepackage{xcolor}
\usepackage{hyperref}

\hypersetup{
  colorlinks=true,
  linkcolor=blue!55!black,
  citecolor=blue!55!black,
  urlcolor=blue!55!black,
  pdftitle={Before Attention Sinks Form: Hidden-State Geometry Reveals and Modulates Their Formation},
  pdfauthor={Fan Yang and Yue Song}
}

\makeatletter
\renewcommand{\normalsize}{\@setfontsize\normalsize{10pt}{11pt}}
\normalsize
\renewcommand{\small}{\@setfontsize\small{9pt}{10pt}}
\renewcommand{\footnotesize}{\@setfontsize\footnotesize{9pt}{10pt}}
\renewcommand{\large}{\@setfontsize\large{12pt}{14pt}}
\renewcommand{\LARGE}{\@setfontsize\LARGE{17pt}{20pt}}
\renewcommand{\section}{\@startsection{section}{1}{\z@}%
  {-2.0ex plus -.5ex minus -.2ex}{1.5ex plus .3ex minus .2ex}%
  {\large\scshape\raggedright}}
\renewcommand{\subsection}{\@startsection{subsection}{2}{\z@}%
  {-1.8ex plus -.5ex minus -.2ex}{.8ex plus .2ex}%
  {\normalsize\scshape\raggedright}}
\renewcommand{\subsubsection}{\@startsection{subsubsection}{3}{\z@}%
  {-1.5ex plus -.5ex minus -.2ex}{.5ex plus .2ex}%
  {\normalsize\scshape\raggedright}}
\renewcommand{\paragraph}{\@startsection{paragraph}{4}{\z@}%
  {1.5ex plus .5ex minus .2ex}{-1em}{\normalsize\bfseries}}
\renewcommand{\maketitle}{%
  \par\begingroup
  \noindent{\LARGE\scshape\@title\par}%
  \vspace{0.13in}%
  \noindent\hspace*{8pt}{\bfseries\@author\par}%
  \vspace{0.28in}%
  \endgroup
}
\renewenvironment{abstract}{%
  \par\vspace{0.075in}
  \begin{center}{\large\scshape Abstract}\end{center}%
  \vspace{0.5ex}%
  \begin{list}{}{\setlength{\leftmargin}{0.5in}%
                 \setlength{\rightmargin}{0.5in}%
                 \setlength{\topsep}{0pt}%
                 \setlength{\partopsep}{0pt}}\item\relax
}{\end{list}\vspace{1ex}}
\makeatother

\title{Before Attention Sinks Form: Hidden-State Geometry Reveals and Modulates Their Formation}
\author{
  \begin{tabular}{@{}l@{}}
    {\normalsize\bfseries Fan Yang, Yue Song\textsuperscript{*}}\\[-1pt]
    {\normalsize\mdseries College of AI, Tsinghua University}
  \end{tabular}
}
    
\date{}

\newcommand{\Afeat}{\mathcal{A}}
\newcommand{\Sfeat}{\mathcal{S}}
\newcommand{\one}{\mathbf{1}}

\newcommand{\appendixcontentline}[3]{%
  \par\noindent
  \makebox[3.4em][l]{#1}%
  \parbox[t]{0.74\linewidth}{#2}%
  \nobreak\leaders\hbox to .45em{\hss.\hss}\hfill\kern0pt
  \nobreak\makebox[2em][r]{#3}\par
}

\begin{document}
\maketitle
\renewcommand{\thefootnote}{*}
\footnotetext{Corresponding author.}
\renewcommand{\thefootnote}{\arabic{footnote}}

\begin{abstract}
Attention sinks are usually studied after they have formed. We ask whether a
measurable hidden-state change appears beforehand and whether changing that
geometry alters what happens later. To remove network depth as a shortcut, we
compare prompts in the same model and at the same layer. When sink timing
varies across prompts, simple spectral summaries of the hidden states add
information beyond current attention, with the clearest prompt-specific signal
in a short window about two to three layers before sink formation. The
predictive strength varies across cohorts, so we treat the spectrum as a readout
of a pre-sink state change rather than a universal warning statistic. The signal
is broader than one measured extreme activation. We then modify only the
singular values of the post-layer hidden states after current attention has
already been computed. These interventions change later sink behavior in
Qwen2.5, GLM-Z1, and GPT-OSS relative to norm-matched random spectral
perturbations. In Qwen2.5, the immediate suppressive effect is accompanied by a
reduction in next-layer query--key preference for position~0, and the learned
spectral directions remain stable when estimated from substantially less
training data. Together, the results show that attention sinks are preceded by
a short spectral reorganization of hidden states, and directly perturbing this
geometry can causally alter subsequent sink formation. The source code for the observational analyses,
spectral interventions, and robustness experiments
is available at
\url{https://github.com/fyangRobert/attention-sink-geometry}
\end{abstract}

\section{Introduction}
\label{sec:introduction}

Attention sinks are a widely observed behavior in transformer-based large
language models (LLMs): a large fraction of attention can concentrate on an
early or special token even when that token is not important to the current
prediction \citep{xiao2023streamingllm}. They also matter for long-context
inference and KV-cache management, where retaining sink tokens can stabilize
generation \citep{xiao2023streamingllm,kvsink2025}. Yet most existing work
studies sinks after they have already appeared. Much less is known about what
happens in the hidden states before a sink forms.

Our main finding is simple:
\textbf{attention sinks are preceded by a short spectral reorganization of
hidden states, and directly perturbing this geometry can causally alter
subsequent sink formation.}
The first part is observational: hidden-state geometry changes before the
attention pattern becomes visible. The second part is causal: modifying the
same geometry after the current layer has already finished computing attention
can change sink behavior in later layers.

Studying the observational part requires a strict control for network depth.
Sink formation is often tied to a characteristic layer, so a model can appear
to predict future sink formation simply by learning where it is in the
network. We therefore compare prompts only within the same model and at the
same current layer. In this setting, model identity and depth are fixed, and
the question is whether prompts that have not yet formed a sink already differ
in their hidden states according to how soon a sink will subsequently appear.

We measure this structure using simple summaries of the singular spectrum of
the prompt-specific token-by-hidden-state matrix. Where sink timing varies
across prompts, these spectral features add information beyond current
attention. The pooled effect extends over several short horizons, but the
clearest prompt-specific evidence under strict within-cell evaluation is
concentrated roughly two to three layers before sink formation. Its predictive
strength varies across prompt cohorts, so we interpret the spectrum as a
readout of a pre-sink representation change rather than as a universal warning
statistic.

\begin{figure}[t]
    \centering
    \includegraphics[width=\linewidth]
    {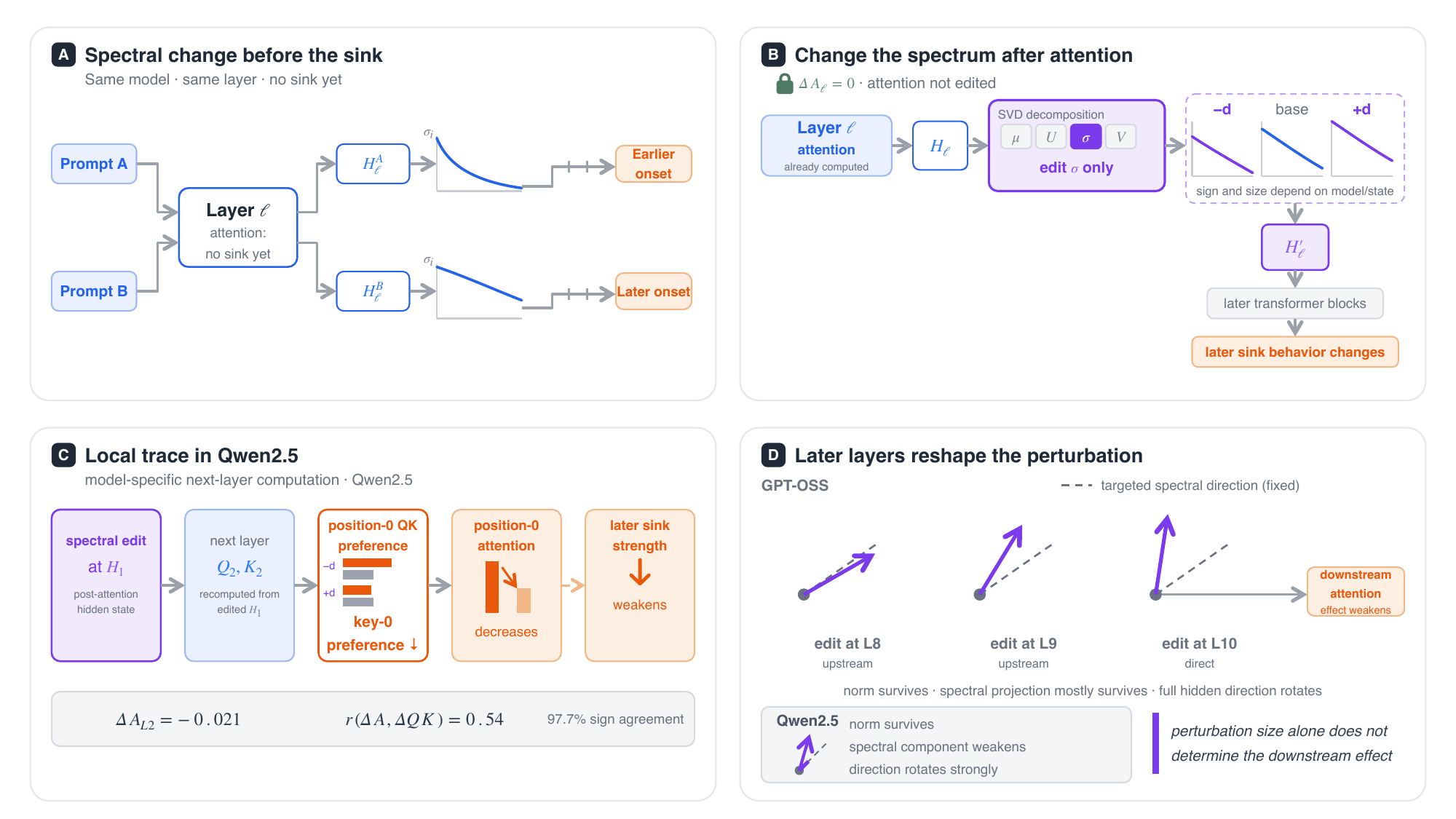}
    \caption{\textbf{Overview of the observational and causal analyses.}
    \textbf{(A)} We compare prompts in the same model and at the same layer
    before either has formed a sink, and ask whether their hidden-state spectra
    already differ before their later sink timing diverges.
    \textbf{(B)} After the source-layer attention has been computed, we modify
    only the singular values of the post-layer hidden state while keeping the
    token mean and singular subspaces fixed; current attention is therefore
    unchanged, and we measure how later sink behavior responds.
    \textbf{(C)} In Qwen2.5, the immediate effect of the spectral edit is
    associated with a change in the next layer's relative QK preference for
    position~0 and a corresponding change in position-0 attention.
    \textbf{(D)} Tracking the perturbation across layers shows that its norm
    can survive while its full hidden-state direction rotates and its targeted
    spectral component may either persist or weaken, helping explain why the
    downstream effect depends on where the edit is introduced.}
    \label{fig:intro-overview}
\end{figure}

\begin{table}[t]
\centering
\caption{
\textbf{Key results and their reference points.}
For observational rows, AP is the prompt-grouped out-of-fold average precision
for predicting future sink formation at the specified horizon. We report the
attention-only baseline $\mathrm{AP}(\Afeat)$ and the gain from adding
hidden-state spectral features,
$\Delta\mathrm{AP}=\mathrm{AP}(\Afeat+\Sfeat)-\mathrm{AP}(\Afeat)$.
Comparisons are made within the same model and current layer.
Intervention rows instead report changes in later sink strength under
controlled post-layer spectral edits and therefore do not use AP.
}
\label{tab:intro-key-results}
\footnotesize
\setlength{\tabcolsep}{2.8pt}
\begin{tabular}{
p{0.18\linewidth}
p{0.28\linewidth}
p{0.31\linewidth}
p{0.17\linewidth}}
\toprule
Question
&
What is compared
&
Main result
&
What it means
\\
\midrule

Does the spectrum add information before a sink forms?
&
Current attention only ($\Afeat$) vs.\ current attention plus spectrum
($\Afeat+\Sfeat$), with model and layer matched.
&
Exact~1:
AP $.857 \rightarrow .913$
($\Delta$AP $=+.056$, $[.028,.087]$).

Strict within-cell:
Within~2 $+.107$ $[.048,.155]$;
Within~3 $+.053$ $[.002,.101]$;
Within~4 $-.021$ $[-.037,.029]$.
&
The extra information is clearest about two to three layers before the sink.
\\

Does changing the spectrum change later sink behavior?
&
For the learned direction, opposite signs ($+d$ vs.\ $-d$) are compared at
the same perturbation size. That directional gap is then tested against gaps
from norm-matched random spectral directions.
&
Later-sink directional gap:

Qwen2.5: $-.00833$, $p=.0308$;

GLM-Z1: $-.01525$, $p=.0308$;

GPT-OSS: $+.05625$, two-sided $p=.0462$;
30/30 fresh prompts have the same positive sign.
&
Yes, but the direction and size of the effect depend on model and state.
\\

What changes immediately after the Qwen2.5 edit?
&
Opposite signs of the learned spectral direction are compared at the same
perturbation size; source-layer attention is unchanged.
&
Next-layer position-0 attention changes by $-.0214$.

The QK-preference change tracks the attention change:
$r=.544$, with 97.7\% sign agreement.
&
The immediate suppressive effect is associated with lower next-layer
preference for position~0.
\\

Can the edit direction be estimated from less data?
&
Directions estimated from reduced training sets are compared with the
full-data direction or with an independently estimated half-sample direction.
&
Qwen2.5:
disjoint-half direction cosine $.979$.

GPT-OSS:
half-data effect $+.05634$
vs.\ full-data effect $+.05625$
(effect ratio $1.002$).
&
The learned direction and its downstream effect are stable under substantial
data reduction.
\\

\bottomrule
\end{tabular}
\end{table}

The spectral change is not well explained by a single unusually large
activation. Compact measures of hidden-state amplitude and concentration leave
substantial incremental spectral information, whereas broader descriptions of
the distributed activation state contain much of the same predictive
information. Qwen3 provides a complementary timing example: 10 prompts form
sinks at layer~3 and 40 at layer~7, yet under a stringent criterion the abrupt
extreme-activation event occurs at layer~6 for both groups. Its timing therefore
cannot explain the prompt-specific difference between early and late sink
formation.

We next intervene directly on the spectral geometry. The edit is applied after
the source layer has already computed attention and changes only the singular
values of the post-layer hidden state, while preserving the token mean,
singular subspaces, and centered spectral energy. Random spectral directions
are matched to the same realized perturbation norm. These spectrum-only edits
change later sink behavior in Qwen2.5, GLM-Z1, and GPT-OSS, while the
source-layer attention remains unchanged. The effect is not universal in sign:
the naturally observed near-state direction suppresses later sink strength in
Qwen2.5 and GLM-Z1, whereas the frozen GPT-OSS direction strengthens its later
full-attention sink endpoint. Thus the spectrum is a causally actionable
coordinate of the hidden state without being a one-dimensional
``toward-sink'' axis.

The downstream effect also depends on what later transformer blocks do to the
perturbation. In Qwen2.5, the immediate suppressive effect is accompanied by a
reduction in the next layer's relative query--key preference for position~0.
Farther downstream, GPT-OSS can preserve both perturbation norm and most of the
targeted spectral component while rotating the full hidden-state displacement
and sharply weakening the attention effect. Qwen2.5 shows a related but
different pattern: the perturbation norm survives, but much of the targeted
spectral component is filtered and the remaining hidden-state change is
strongly reoriented. These results help explain why the effect of a spectral
edit depends on where it is introduced.

\paragraph{Contributions.}
We make three contributions:
\begin{itemize}[leftmargin=*,nosep]
    \item \textbf{Pre-sink state change.}
    We identify a short spectral reorganization before sink formation using
    same-model, same-layer comparisons, and characterize the cohort dependence
    and boundary conditions of this observational signal.

    \item \textbf{Controlled spectral intervention.}
    We show that modifying only post-layer singular values can change later
    sink behavior while current attention remains fixed, with evidence from
    Qwen2.5, GLM-Z1, and a frozen GPT-OSS confirmation against norm-matched
    random directions.

    \item \textbf{Local and cross-layer explanation.}
    We trace the immediate Qwen2.5 effect to reduced next-layer position-0
    query--key preference and show that later transformer blocks reshape the
    perturbation, explaining why its downstream effect depends on source layer
    and model state.
\end{itemize}

\section{Related Work}
\label{sec:related-work}

\paragraph{Attention sinks.}
StreamingLLM showed that keeping early sink tokens can stabilize streaming
generation \citep{xiao2023streamingllm}. Later work studied where sinks appear,
how training, normalization, context length, and data distribution affect them,
and how sink-aware methods can improve KV-cache management
\citep{yu2024hidden,gu2024whenattentionsink,barbero2025firsttoken,kvsink2025}.
Mechanistic work has related sinks to active--dormant attention heads and the
joint emergence of value-state drains and residual-state peaks
\citep{guo2024activedormant}, while a recent geometric account interprets sink
tokens as reference frames that anchor representational spaces
\citep{ruscio2025geometry}. Large residual activations have also been linked to
sink behavior \citep{sun2024massive}, while representation compression has been
associated with related sink phenomena \citep{queipo2025compression}. We
instead ask what changes \emph{within a forward pass before} the sink appears
for a particular prompt and whether directly changing that hidden-state
geometry alters later sink formation.

\paragraph{Hidden-state geometry.}
Transformer representations vary systematically across layers in anisotropy,
dominant directions, rank, clustering, and signal propagation
\citep{ethayarajh2019contextual,timkey2021rogue,dong2021attention_rank_collapse,
noci2022signal,kedia2024transformersstable,geshkovski2023clusters}. Recent work
connects layerwise and spectral geometry to attention and representation
dynamics \citep{viswanathan2025geometry,li2026dispersion,cancedda2024spectral},
and spectral transformations have been used to remove information from neural
representations \citep{shao2023gold}. Cancedda's spectral filters partition the
singular directions of the model's vocabulary embedding and unembedding
matrices; our object is different: the singular spectrum of the
prompt-specific token-by-hidden-state matrix at each layer. We use this spectrum
first to measure a short pre-sink change and then to test that geometry
directly.

\paragraph{Interventions in hidden representations.}
Causal tracing and activation patching measure downstream effects of replacing
or restoring hidden states \citep{meng2022locating}; related work steers models
with learned activation directions \citep{li2023inference,turner2023steering,
zou2023representation,lai2025joint}. Our intervention keeps the token mean and
singular subspaces fixed, changes only the singular values, and compares the
result with random spectral perturbations of the same size.

Taken together, prior work explains how attention sinks behave after they
appear and provides general tools for intervening on hidden representations.
What remains unclear is whether a prompt-specific representation change appears
\emph{before} a sink forms and whether directly changing that geometry alters
later sink behavior. These are the two questions we test below.

\section{Measuring the Pre-Sink Change}
\label{sec:measurement}

This section defines the two tests used throughout the paper. First, we ask
whether the current hidden state contains prompt-specific information about
future sink formation while model and current layer are held fixed. Second, we
define a post-layer spectrum-only edit that leaves current attention unchanged,
allowing later changes to be tested directly. We use later layers only to
determine when the sink forms. The first two subsections define the
observational test; the last defines the intervention.

\subsection{Defining when a sink forms}
\label{sec:sink-onset}

For prompt $i$, let $A^{(i,\ell)}_{a,q,k}$ be the attention weight from query
$q$ to key $k$ in head $a$. We focus on position $k=0$ and average over late
queries,
\begin{equation}
\bar a^{(i,\ell)}_a(q_{\min})=
\frac{1}{T_i-q_{\min}}\sum_{q=q_{\min}}^{T_i-1}A^{(i,\ell)}_{a,q,0}.
\end{equation}
The fraction of heads whose average exceeds $\varepsilon$ is
\begin{equation}
s^{(i,\ell)}_{\varepsilon}(q_{\min})=
\frac{1}{N_h}\sum_a\one\!\left\{\bar a^{(i,\ell)}_a(q_{\min})>\varepsilon\right\}.
\end{equation}
With $q_{\min}=16$, $\varepsilon=0.3$, and $\theta=0.5$, the primary sink-
formation layer is
\begin{equation}
\tau_i=\min\{\ell:s^{(i,\ell)}_{\varepsilon}(q_{\min})\ge\theta\}.
\end{equation}
We study the canonical position-0 sink throughout; whether the same precursor
appears before sinks at other token positions is outside the scope of this
work. We analyze only $\ell<\tau_i$. Because a sink-like state does not always remain
present after its first appearance, we also repeat the analysis with definitions
requiring the condition to hold for two or three consecutive layers.

To measure how early the hidden-state change can be detected, we use
\begin{equation}
y^{\rm exact}_{i,\ell,h}=\one\{\tau_i-\ell=h\},\qquad
y^{\rm within}_{i,\ell,h}=\one\{1\le\tau_i-\ell\le h\},
\end{equation}
for $h\in\{1,2,3,4,6\}$. The first asks whether the sink forms exactly $h$
layers later; the second asks whether it forms within the next $h$ layers.
These are analysis labels rather than a proposed deployed warning system.
Threshold and context-length checks are in
Appendix~\ref{app:formation-robustness}.

\subsection{Hidden-state spectrum and same-layer evaluation}
\label{sec:spectral-state}
\label{sec:matched-evaluation}

Let $H_i^{(\ell)}\in\mathbb{R}^{T_i\times d}$ be the hidden-state matrix after
layer $\ell$. We center it across tokens,
\begin{equation}
\widetilde H_i^{(\ell)}=H_i^{(\ell)}-\mathbf{1}\bar h_i^{(\ell)\top},
\end{equation}
and compute its singular values. The spectral family $\Sfeat$ contains five
summaries---log--log slope, spectral entropy, effective rank, leading singular
energy, and top-five energy---together with nearby layerwise changes. These
statistics describe how variation across token representations is distributed
across hidden directions. Full definitions are in Appendix~\ref{app:features}.

For each target, we keep only model--layer groups containing both prompts whose
sinks form soon and prompts whose sinks form later. Within such a group, model
and current depth are identical. We compare an attention feature family
$\Afeat$ with $\Afeat+\Sfeat$ using $\ell_2$-regularized logistic models. All
rows from one prompt stay in the same fold; we use 5-fold grouped outer
evaluation, 4-fold grouped inner model selection, fold-local preprocessing, and
prompt-level bootstrap uncertainty. Average precision (AP) is the primary
metric. Stricter tests remove model--layer spectral offsets or evaluate AP
directly inside each matched group.

\noindent\textbf{Evaluation Metric.}
For a target horizon $h$, each pre-sink observation receives a binary label:
positive if a sink forms at the specified future horizon (Exact-$h$), or
within the next $h$ layers (Within-$h$), and negative otherwise. We use
average precision (AP) to measure how well the current state ranks positive
prompts above negative prompts. Our primary quantity is the incremental
out-of-fold performance
\begin{equation}
\Delta\mathrm{AP}
=
\mathrm{AP}(\Afeat+\Sfeat)
-
\mathrm{AP}(\Afeat),
\end{equation}
where $\Afeat$ is the current-attention baseline and $\Sfeat$ is the
hidden-state spectral feature family. Thus $\Delta\mathrm{AP}>0$ means that
the hidden-state spectrum adds information about whether a sink will form soon
beyond what is already available from current attention. Unless otherwise
stated, AP is computed from prompt-grouped out-of-fold predictions in the
matched model--layer cohort; stricter within-cell analyses evaluate the same
ranking question separately inside each matched group.

\noindent\textbf{Primary result and controls.}
For the primary Exact~1 target, the attention baseline achieves
$\mathrm{AP}=.857$, while adding spectral features increases performance to
$.913$, giving $\Delta\mathrm{AP}=+.056$ with a prompt-bootstrap 95\%
confidence interval of $[.028,.087]$. The increment remains positive after
all threshold-derived sink-score features are removed from the attention
baseline ($\Delta\mathrm{AP}=+.085$ $[.051,.123]$), showing that the result is
not driven by reusing threshold-based quantities that also define when the
sink first appears. Across future horizons, the pooled association extends
through roughly four layers, while strict within-cell and sustained-onset
analyses place the clearest prompt-specific signal about two to three layers
before the sink forms.

\subsection{Direct spectral intervention}
\label{sec:causal-method}

Prediction alone does not show that the spectral change matters. For the causal
experiment, write the centered post-layer state as
\begin{equation}
\widetilde H=H-\mathbf{1}\mu^\top=U\,\mathrm{diag}(\sigma)\,V^\top.
\end{equation}
Using training data only, we estimate a direction $d$ in normalized
log-spectrum space from naturally occurring state differences. At test time we
modify the leading 32 singular values (or all available nonzero values),
\begin{equation}
\log\sigma'=\log\sigma+\eta d,
\end{equation}
rescale them to preserve centered Frobenius energy, and reconstruct
\begin{equation}
H'=\mathbf{1}\mu^\top+U\,\mathrm{diag}(\sigma')\,V^\top.
\end{equation}
The token mean and singular subspaces remain fixed. For each prompt and
direction, $\eta$ is chosen so that
\begin{equation}
\rho=\frac{\|H'-H\|_F}{\|H\|_F}
\end{equation}
matches a fixed budget. Random spectral directions are calibrated to the same
$\rho$, controlling for the effect of an equally large but untargeted change.

The edit is applied after layer $\ell$ has already computed its attention.
Therefore
\begin{equation}
A'_\ell=A_\ell,
\end{equation}
and the experiment asks whether later attention changes systematically. Every
causal run verifies zero change in the source-layer sink statistic. Complete
construction and random-direction controls are in
Appendix~\ref{app:causal-details}.

\section{The Pre-Sink Spectral Change}
\label{sec:warning}
\label{sec:formation-regimes}
\label{sec:interpretation}
\label{sec:distributed-state}

This section tests the first half of our main claim: whether spectral geometry
changes before a sink forms and how close to onset that change is most clearly
visible.

The same-layer comparison is meaningful only when a model has both a measurable
pre-sink interval and prompt-to-prompt variation in sink timing. Qwen2.5,
Qwen3, GLM-4, and GLM-Z1 provide usable support in the primary setting; the
full support table and model-specific counts are in
Appendix~\ref{app:model-regimes}. Figure~\ref{fig:transition-window} gives the
main temporal result.

\begin{figure}[t]
\centering
\includegraphics[width=\linewidth]
{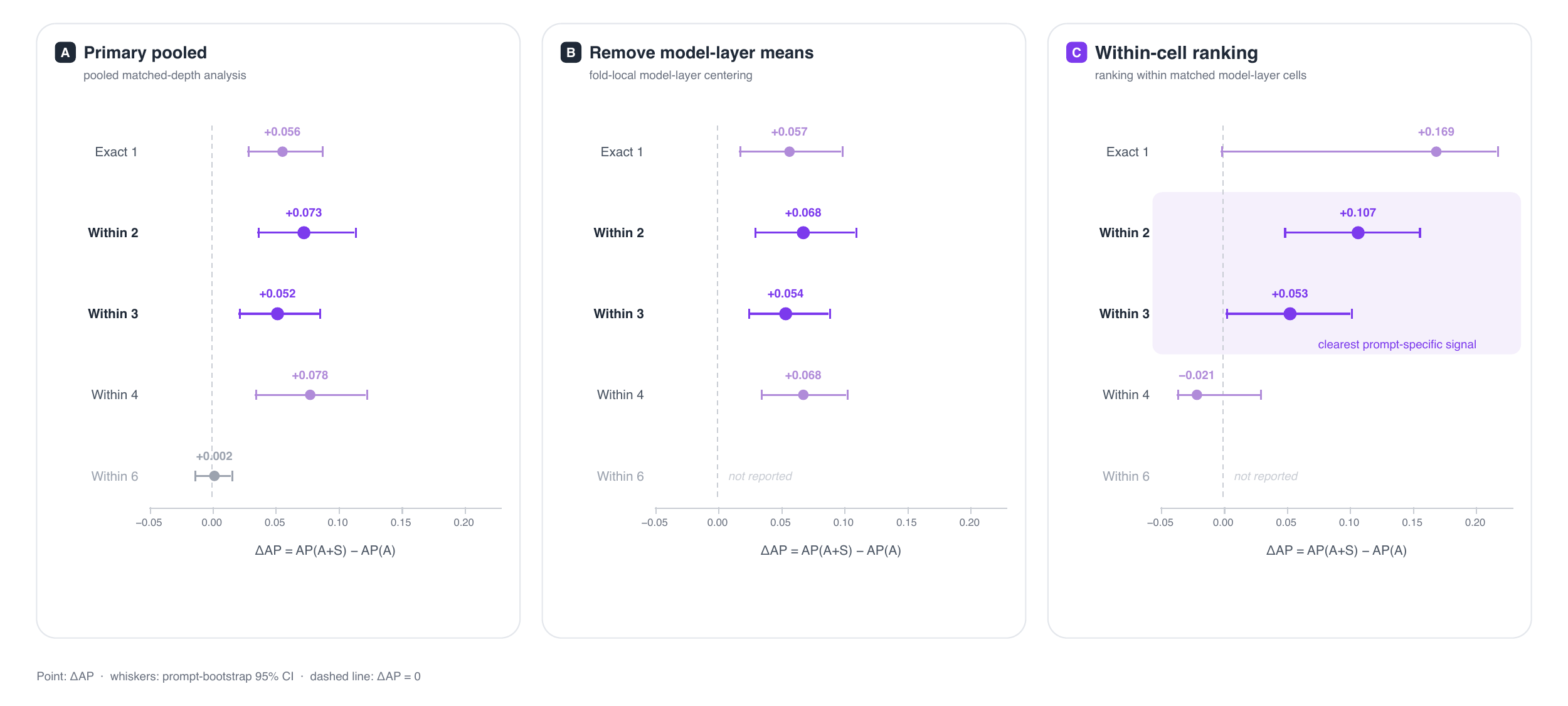}
\caption{\textbf{The spectral increment is concentrated in a short pre-sink
window.}
\textbf{(A)} In the primary pooled matched-depth analysis, adding spectral
features to current attention improves prediction over several short horizons
but not at six layers.
\textbf{(B)} The pattern remains after removing model--layer spectral means
using training-fold statistics.
\textbf{(C)} When ranking is evaluated entirely within matched model--layer
cells, the clearest prompt-specific gains occur at Within~2 and Within~3,
while the Within~4 estimate is near zero.
Points show $\Delta$AP and horizontal bars show prompt-bootstrap $95\%$
confidence intervals; the vertical dashed line marks $\Delta$AP$=0$.}
\label{fig:transition-window}
\end{figure}

The primary pooled Exact~1 comparison improves from
$\mathrm{AP}(\Afeat)=.857$ to $\mathrm{AP}(\Afeat+\Sfeat)=.913$, giving
$\Delta\mathrm{AP}=+.056$ with 95\% CI $[.028,.087]$. The stricter
within-cell analysis is more informative about timing: Within~2 gives
$+.107$ $[.048,.155]$, Within~3 gives $+.053$ $[.002,.101]$, and
Within~4 is near zero. Table~\ref{tab:main-observational-summary} collects the
few additional results needed to interpret this pattern; the full horizon
grids and sensitivity analyses are deferred to Appendix~\ref{app:warning-full}.

\begin{table}[t]
\centering
\caption{\textbf{Main observational results.}
Each row states the comparison before the spectral gain is reported. The strict
same-model, same-layer result localizes the clearest prompt-specific signal to
the next two to three layers.}
\label{tab:main-observational-summary}
\footnotesize
\setlength{\tabcolsep}{2.6pt}
\begin{tabular}{
p{0.19\linewidth}
p{0.25\linewidth}
p{0.30\linewidth}
p{0.18\linewidth}}
\toprule
Question & What is compared & Main result & Takeaway \\
\midrule

Does the spectrum add information?
&
Attention only vs.\ attention + spectrum, with model and layer matched.
&
Exact~1:
AP $.857 \rightarrow .913$;
$\Delta$AP $=+.056$ $[.028,.087]$.
&
Spectrum adds information beyond current attention.
\\

When is the clearest prompt-specific signal?
&
Ranking is evaluated only among prompts in the same model and current layer.
&
Within~2:
$+.107$ $[.048,.155]$;

Within~3:
$+.053$ $[.002,.101]$;

Within~4:
$-.021$ $[-.037,.029]$.
&
The clearest signal is about two to three layers before the sink.
\\

Does the result depend on threshold-derived attention features?
&
Continuous attention only vs.\ continuous attention + spectrum.
&
Exact~1:
AP $.833 \rightarrow .918$;

Within~2:
$.793 \rightarrow .889$;

Within~6:
$.899 \rightarrow .912$, with CI including zero.
&
The short-range result does not come from reusing the onset threshold.
\\

Is the effect size stable across prompt cohorts?
&
The same frozen Qwen3 analysis is evaluated on the expanded cohort and on the
unseen post-audit subset.
&
$N=500$, Exact~2:
$\Delta$AP $=+.232$ $[.117,.340]$;

unseen $N=372$:
$+.102$ $[-.035,.223]$.
&
The short transition persists as a pattern, but predictive strength is cohort
dependent.
\\

\bottomrule
\end{tabular}
\end{table}

The robustness checks support a hierarchy of claims rather than a universal
horizon. Removing model--layer spectral means preserves the pooled short-range
increment, while ranking entirely within matched model--layer cells localizes
the clearest prompt-specific evidence to Within~2 and Within~3. The same
localization survives a two-block sustained onset definition. Equal model
weighting, leave-one-model-out refits, nonlinear attention baselines, and a
prompt-shuffling null provide complementary checks. Importantly, the expanded
Qwen3 cohort does not justify a universal two-layer predictor: after removing
the 128 prompts examined in an earlier strengthening audit, the unseen
372-prompt Exact~2 estimate remains positive but its interval includes zero.
We therefore emphasize the short representation change, not a fixed predictive
effect size. Full numerical results and the overlap audit are in
Appendix~\ref{app:warning-full}.

\subsection{What does the spectral change reflect?}

The spectrum is not well described as a detector of one exceptional
activation. Compact hidden-state amplitude and concentration controls leave a
substantial positive spectral increment, whereas broader descriptions of the
distributed activation state reduce the remaining increment toward zero.
Capacity-matched nuisance, subset, random-projection, and prompt-alignment
controls show that this overlap requires genuine prompt-specific activation
information but is not unique to one handcrafted basis. A complementary Qwen3
timing audit reaches the same qualitative conclusion from a different angle:
10 prompts form a sink at block~3 and 40 at block~7, while the stringent
extreme-activation event occurs at block~6 for both groups across thresholds
from $2\times10^3$ to $2\times10^4$. Thus the measured extreme-event depth
cannot explain the prompt-specific difference in sink timing. We interpret the
spectrum as a compact readout of a broader hidden-state reorganization, not as
a complete causal mechanism. The full conditioning, capacity-control, and
timing analyses are in Appendix~\ref{app:activation-state-redundancy}.

Taken together, the observational evidence supports a short,
prompt-specific representation change rather than a universal predictor or a
detector of one extreme activation. These analyses show when the spectral
change appears and what information it overlaps with, but they do not show
that the geometry matters for later computation. We therefore turn next to a
direct intervention on the spectrum.

\section{Causal Effects of Spectral Intervention}
\label{sec:causal-results}
\label{sec:gptoss-causal}
\label{sec:cross-model-causal}
\label{sec:direction-stability}
\label{sec:transport}

Prediction alone does not show that the pre-sink geometry matters. We therefore
edit only the singular values of the post-layer hidden state after the source
layer has already computed attention, while preserving the token mean,
singular subspaces, and centered spectral energy. Random spectral directions
are calibrated to the same realized hidden-state norm, and every reported run
verifies zero change in the source-layer sink statistic.
Table~\ref{tab:causal-main} summarizes the main intervention, local-trace,
stability, and propagation results; full random-direction tests, dose
responses, diagnostics, and construction details are in
Appendix~\ref{app:causal-details}.

\begin{table}[t]
\centering
\caption{\textbf{Main causal and cross-layer results.}
Intervention effects are directional contrasts between opposite signs of the
learned spectral direction at matched perturbation size; their empirical
$p$ values are computed against norm-matched random spectral directions.}
\label{tab:causal-main}
\footnotesize
\setlength{\tabcolsep}{2.5pt}
\begin{tabular}{
p{0.18\linewidth}
p{0.25\linewidth}
p{0.30\linewidth}
p{0.19\linewidth}}
\toprule
Test & What is compared & Main result & Takeaway \\
\midrule

Qwen2.5 intervention
&
Opposite signs of the learned direction at $\rho=.05$; the resulting gap is
tested against matched random directions.
&
Future sink-strength gap
$=-.00833$;
empirical $p=.0308$.
&
Changing the spectrum suppresses later sink strength in this setting.
\\

GLM-Z1 intervention
&
Opposite signs of the learned direction at $\rho=.025$; the same matched-random
test is used.
&
Future sink-strength gap
$=-.01525$;
empirical $p=.0308$.
&
Independent cross-model support; the sign is again suppressive.
\\

GPT-OSS frozen test
&
Frozen L10 learned direction at $\rho=.025$; later sink strength is averaged
over full-attention layers 11, 13, and 15.
&
Gap $=+.05625$
$[.04983,.06267]$;
two-sided $p=.0462$;
30/30 prompts positive.
&
Fresh-cohort confirmation with the opposite sign.
\\

Qwen2.5 next-layer trace
&
Opposite signs of the learned direction at the same budget, inspected at the
next attention layer.
&
Position-0 attention:
$-.0214$;

QK--attention correlation:
$r=.544$;

97.7\% sign agreement.
&
The immediate suppressive effect tracks lower position-0 QK preference.
\\

Direction stability
&
Directions estimated from reduced data vs.\ full-data or independent
half-sample estimates.
&
Qwen2.5 disjoint-half cosine $.979$;

GPT-OSS:
half-data effect $.05634$
vs.\ full-data $.05625$.
&
The learned direction is not fragile to training-set reduction.
\\

Cross-layer change
&
Upstream edits propagated to the same hidden state vs.\ a direct local edit.
&
GPT-OSS:
96--100\% spectral projection retained,
hidden cosine $.65$--$.81$;

Qwen2.5:
24\% retained,
cosine $.18$.
&
Later blocks can preserve perturbation size while changing its direction and
downstream effect.
\\

\bottomrule
\end{tabular}
\end{table}

Across Qwen2.5, GLM-Z1, and GPT-OSS, the supported causal statement is
deliberately limited: \textbf{changing the singular spectrum after a layer has
finished computing can change sink behavior in later layers while leaving the
current layer's attention unchanged.} Qwen2.5 provides a more local view of
the first downstream attention computation: the suppressive edit reduces
position-0 attention at layer~2, and this change closely tracks a reduction in
relative position-0 query--key preference. This is a model-specific
computational trace, not evidence that the same QK route or sign must hold
across architectures.

The cross-layer measurements explain why the same nominal edit can have
different effects at different source layers. In GPT-OSS, upstream
perturbations reach the local layer with norms comparable to or larger than the
direct edit and retain roughly 96--100\% of the targeted spectral projection,
yet the full hidden-state displacement rotates substantially and the downstream
effect becomes much weaker. Qwen2.5 shows a related but different pattern: the
norm survives, but only about 24\% of the targeted spectral projection remains
and the hidden displacement is strongly reoriented. Transformer blocks
therefore reshape the intervention; perturbation size alone does not determine
its downstream effect. Full propagation measurements are in
Appendix~\ref{app:causal-details}.

Together, these results show that changing the spectrum can change later
attention, while subsequent transformer blocks determine how strongly and in
what direction that change is expressed.

\section{Scope and Boundary Conditions}
\label{sec:robustness}

The matched-depth analysis requires a measurable pre-sink interval and
prompt-to-prompt variation in onset depth; the causal direction rule likewise
requires its pre-specified support groups to exist. Table~\ref{tab:main-boundary-summary}
summarizes the main settings in which one of these requirements changes or
fails.

\begin{table}[t]
\centering
\caption{\textbf{Main boundary conditions.} Lack of an estimable comparison is
not treated as negative evidence for the underlying spectral phenomenon.}
\label{tab:main-boundary-summary}
\small
\setlength{\tabcolsep}{3.2pt}
\begin{tabular}{p{0.22\linewidth}p{0.35\linewidth}p{0.35\linewidth}}
\toprule
Setting & Boundary & Consequence \\
\midrule
Gemma / Llama
&
Sink appears very early or is present at the first measured layer.
&
Little or no ordinary pre-sink interval is available.
\\

Qwen3, $T=256$
&
499/500 prompts first cross at layer~7.
&
Prompt-specific matched-depth timing is effectively non-identifiable.
\\

Qwen2.5, $T=256$
&
All 429 crossing prompts first cross at layer~4.
&
No onset-depth variation exists among crossing prompts.
\\

GLM-Z1, $T=256$
&
Onset remains spread across several layers.
&
The strict same-layer analysis remains estimable and again localizes to
Within~2--3.
\\

Fresh GLM-4 / Qwen3 causal preparation
&
The pre-specified near/far support required by the frozen direction rule is
absent.
&
The planned comparison is not replaced post hoc with a different grouping.
\\

GPT-OSS / Qwen3.6
&
Position-0 visibility depends on attention type / sink-like states can
disappear and reappear.
&
The ordinary single persistent-onset target is not architecture invariant.
\\
\bottomrule
\end{tabular}
\end{table}

These cases separate absence of evidence from absence of a measurable target.
Longer context can collapse the prompt variation needed by the same-layer
analysis, while attention architecture can change what ``sink formation''
means. GPT-OSS is therefore evaluated at later full-attention checkpoints, and
Qwen3.6 is treated as a recurrent sink-like process rather than an ordinary
persistent-onset model. Complete threshold, context-length, support, and
architecture-specific audits are in Appendices~\ref{app:formation-robustness}
and~\ref{app:architecture-boundaries}.

\section{Discussion and Limitations}
\label{sec:discussion}

The evidence supports two linked claims. First, attention sinks are preceded by
a short spectral reorganization of hidden states. Same-layer comparisons remove
network depth as a shortcut, with the clearest prompt-specific evidence about
two to three layers before sink formation. The expanded Qwen3 cohort shows that
the effect can be strong, but the post-audit holdout shows that its magnitude is
cohort dependent; we therefore interpret it as a short representation change
rather than a universal onset predictor.

Second, directly changing this geometry can alter later sink behavior.
Post-layer edits leave current attention fixed and separate from norm-matched
random spectral perturbations in Qwen2.5, GLM-Z1, and GPT-OSS. The frozen
GPT-OSS confirmation and half-data direction tests further show that the effect
is not tied to one fragile direction estimate. These experiments establish
functional dependence, not a practical control policy.

The interventions also show why no universal direction or local mechanism
should be expected. In Qwen2.5, the near-state direction is suppressive and is
associated with lower next-layer position-0 query--key preference. Farther
downstream, GPT-OSS and Qwen2.5 reshape the perturbation differently: its norm
can survive while the full hidden-state direction rotates and the targeted
spectral component persists or is filtered. Thus the two--three-layer window is
an observational localization, not a universal point of maximal causal
sensitivity.

Several limitations remain. We study only the canonical position-0 sink;
support is uneven across models and prompt cohorts, and longer context can make
the matched-depth task non-identifiable. Singular-value edits may also leave
naturally visited states; norm-matched controls, KL checks, direction-stability
tests, and source-attention invariance reduce but do not eliminate this concern.
Larger perturbations can cause broader functional changes, and GPT-OSS and
Qwen3.6 show that a single persistent sink-onset definition is not architecture
invariant.

\section{Conclusion}

Attention sinks are preceded by a short spectral reorganization of hidden
states, with the clearest same-layer evidence in a roughly two--three-layer
window but cohort-dependent predictive strength. Directly changing the
post-layer spectrum can alter sink formation in later layers while current
attention remains fixed. In Qwen2.5, the immediate suppressive effect is
accompanied by reduced next-layer query--key preference for position~0; farther
downstream, transformer layers can rotate or weaken the perturbation. The
spectrum therefore captures an important part of the pre-sink state without
being a complete description of what determines the sink.

\subsection*{Reproducibility Statement}
All observational analyses keep prompts grouped across folds, fit preprocessing
and model selection on training data only, and use later layers only to define
when a sink subsequently forms. Prompt hashes are used to audit overlap between
the original, expanded, and post-audit holdout cohorts. Intervention directions
are estimated from training data only; natural and random spectral edits are
norm matched, applied after source-layer attention is computed, and audited for
zero source-layer sink change. GPT-OSS freezes its source layer, direction,
perturbation size, and future full-attention outcome before evaluation on the
separate 30-prompt cohort. The appendix and supplement report full feature
definitions, support counts, alternative sink definitions, direction-stability
splits, random-direction tests, hashes, seeds, KL/top-1 diagnostics, and
cross-layer measurements. Pre-specified comparisons without sufficient support
are not replaced by post-hoc groupings.

\subsection*{AI Use Statement}
In this work, we used generative AI tools for limited assistance with code
development, LaTeX assistance, and language editing. Their use was restricted
to improving code implementation and the clarity, grammar, and readability of
the manuscript; they were not used to generate, alter, or determine the
reported experimental results. All AI-assisted text and code were reviewed,
tested, and verified by the authors. The authors retain full responsibility
for the accuracy, validity, and integrity of the final manuscript, including
its text, claims, experimental results, code, and accompanying artifacts.

\bibliography{refs}
\bibliographystyle{plainnat}

\appendix

\clearpage
\phantomsection
\section*{Appendix Contents}
\begingroup
\small
\setlength{\parindent}{0pt}
\appendixcontentline{\textbf{A}}{\textbf{Supplementary Methods}}{\textbf{\pageref{app:details}}}
\appendixcontentline{A.1}{Feature definitions}{\pageref{app:features}}
\appendixcontentline{A.2}{Same-layer evaluation and implementation checks}{\pageref{app:same-layer-implementation}}
\appendixcontentline{A.3}{Model-level sink patterns used for the main comparison}{\pageref{app:model-regimes}}
\medskip
\appendixcontentline{\textbf{B}}{\textbf{Additional Evidence for the Pre-Sink Spectral Change}}{\textbf{\pageref{app:warning-full}}}
\appendixcontentline{B.1}{Full horizon results}{\pageref{app:full-horizon-results}}
\appendixcontentline{B.2}{Expanded Qwen3 cohort and post-audit holdout}{\pageref{app:qwen3-expanded}}
\appendixcontentline{B.3}{Removing threshold-derived attention features}{\pageref{app:continuous-only}}
\appendixcontentline{B.4}{How much support comes from each model?}{\pageref{app:matched-support}}
\appendixcontentline{B.5}{Removing residual model--layer differences}{\pageref{app:remove-residual-model-layer}}
\appendixcontentline{B.6}{Alternative sink definitions and equal model weighting}{\pageref{app:sustained-onset}}
\appendixcontentline{B.7}{Nonlinear attention baselines and a prompt-shuffling test}{\pageref{app:nonlinear-baselines}}
\appendixcontentline{B.8}{Leave-one-model-out sensitivity}{\pageref{app:lomo-subsec}}
\appendixcontentline{B.9}{Generalization across prompt families}{\pageref{app:family-transfer}}
\medskip
\appendixcontentline{\textbf{C}}{\textbf{Robustness to Sink Definition and Context Length}}{\textbf{\pageref{app:formation-robustness}}}
\appendixcontentline{C.1}{Changing the sink threshold and query window}{\pageref{app:threshold-query-window}}
\appendixcontentline{C.2}{What happens at longer context lengths?}{\pageref{app:longer-context}}
\medskip
\appendixcontentline{\textbf{D}}{\textbf{Additional Tests of the Extreme-Activation Explanation}}{\textbf{\pageref{app:activation-state-redundancy}}}
\appendixcontentline{D.1}{Conditioning on other activation measurements}{\pageref{app:conditioning-activations}}
\appendixcontentline{D.2}{Is the broader activation summary special?}{\pageref{app:broader-activation-special}}
\appendixcontentline{D.3}{Qwen3: the extreme event is depth-locked across onset groups}{\pageref{app:qwen3-extreme-event}}
\medskip
\appendixcontentline{\textbf{E}}{\textbf{Additional Intervention and Cross-Layer Results}}{\textbf{\pageref{app:causal-details}}}
\appendixcontentline{E.1}{Intervention construction}{\pageref{app:intervention-construction}}
\appendixcontentline{E.2}{Qwen2.5 and GLM-Z1}{\pageref{app:qwen25-glmz1}}
\appendixcontentline{E.3}{GPT-OSS frozen confirmation}{\pageref{app:gptoss-frozen}}
\appendixcontentline{E.4}{Local Qwen2.5 query--key trace}{\pageref{app:qwen25-qk}}
\appendixcontentline{E.5}{Direction-estimation robustness}{\pageref{app:direction-stability}}
\appendixcontentline{E.6}{How the perturbation changes across layers}{\pageref{app:perturbation-transport}}
\medskip
\appendixcontentline{\textbf{F}}{\textbf{Architecture-Specific Cases}}{\textbf{\pageref{app:architecture-boundaries}}}
\appendixcontentline{F.1}{Why GPT-OSS needs a different sink measure}{\pageref{app:gptoss-different-sink}}
\appendixcontentline{F.2}{Why Qwen3.6 does not have one persistent sink-formation layer}{\pageref{app:qwen36-persistent-sink}}
\medskip
\appendixcontentline{\textbf{G}}{\textbf{Exploratory Jacobian Analysis}}{\textbf{\pageref{app:routes}}}
\appendixcontentline{\textbf{H}}{\textbf{Summary of Claims and Evidence}}{\textbf{\pageref{app:scope}}}
\endgroup
\clearpage


\section{Supplementary Methods}
\label{app:details}

This section gives the full feature definitions and evaluation procedures behind the main experiments. Later appendix sections report the complete statistical results, robustness checks, intervention experiments, cross-layer measurements, and architecture-specific cases.

\subsection{Feature definitions}
\label{app:features}

This subsection gives the feature definitions used to measure the pre-sink
transition and to test what information that transition summarizes. Let $L$
denote the number of measured blocks and let
$\bar a^{(i,\ell)}_a$ denote the per-head late-query attention defined in
Section~\ref{sec:sink-onset}. The normalized depth is

\begin{equation}
d_\ell=\frac{\ell}{\max(L-1,1)}.
\end{equation}

Define

\begin{equation}
\alpha^{(i,\ell)}
=
\frac{1}{N_h}
\sum_a
\bar a^{(i,\ell)}_a,
\qquad
\nu_i(q_{\min})
=
\frac{1}{T_i-q_{\min}}
\sum_{q=q_{\min}}^{T_i-1}
\frac{1}{q+1},
\end{equation}

where $\nu_i$ is the attention that position~0 would receive under uniform attention over all causally available keys. The attention enrichment and log enrichment are

\begin{equation}
E^{(i,\ell)}
=
\frac{\alpha^{(i,\ell)}}{\nu_i},
\qquad
L^{(i,\ell)}
=
\log
\frac{\max(\alpha^{(i,\ell)},10^{-12})}
{\max(\nu_i,10^{-12})}.
\end{equation}

The attention family $\Afeat$ consists of $d_\ell$, the mean, standard
deviation, and maximum across the headwise values
$\bar a_a^{(i,\ell)}$, $\nu_i$, $E^{(i,\ell)}$,
$L^{(i,\ell)}$, and sink-head fractions
$s_{\varepsilon}^{(i,\ell)}$ for
$\varepsilon\in\{0.2,0.3,0.4,0.5\}$. For

\begin{equation}
x\in
\{
\alpha,E,L,s_{0.2},s_{0.3},s_{0.4},s_{0.5}
\},
\end{equation}

we additionally include the backward difference

\begin{equation}
\Delta^-x_{\ell}
=
x_{\ell}-x_{\ell-1},
\end{equation}

with the first measured block initialized to zero.

For the spectral family, singular values are restricted to finite values and
clamped below at $10^{-8}$ before fitting. For each statistic

\begin{equation}
z\in
\{
\alpha_{\rm spec},
r_{\rm eff},
H_{\rm spec},
p_1,
E_5
\},
\end{equation}

where

\begin{equation}
E_5
=
\frac{\sum_{j\le5}\sigma_j^2}
{\sum_j\sigma_j^2},
\end{equation}

the spectral family $\Sfeat$ contains the post-block value
$z_{\ell}^{\rm post}$, the current-block shift

\begin{equation}
\Delta^{\rm block}z_{\ell}
=
z_{\ell}^{\rm post}
-
z_{\ell}^{\rm pre},
\end{equation}

and the trailing shift

\begin{equation}
\Delta^{\rm trail}z_{\ell}
=
z_{\ell}^{\rm post}
-
z_{\ell-1}^{\rm post},
\end{equation}

with the first trailing shift initialized to zero. Thus every feature assigned
to block $\ell$ is available no later than the completion of that block.

The activation-state families used to interpret the pre-sink transition are
summarized in Table~\ref{tab:feature-families}. Activation statistics are
computed over valid-token entries. For a branch tensor $Z$, ``max/RMS''
denotes

\begin{equation}
\frac{\max |Z|}
{\sqrt{\operatorname{mean}(Z^2)}},
\end{equation}

and branch sink-norm ratios denote the sink-token $\ell_2$ norm divided by
the mean $\ell_2$ norm over the remaining valid tokens. These families range
from compact measurements of a single extreme activation to broader
descriptions of distributed activation structure.

\begin{table}[htbp]
\centering
\caption{
Activation-state families used to interpret the pre-sink transition.
The distributed family contains 18 amplitude/outlier variables and is
intentionally broader than the compact hidden-state controls.
}
\label{tab:feature-families}
\scriptsize
\setlength{\tabcolsep}{3.0pt}
\begin{tabular}{p{0.19\linewidth}p{0.74\linewidth}}
\toprule
Family & Variables / operational definition \\
\midrule

QK preference
&
Sink log-probability mean/max, sink-versus-best-competitor
log-probability margin mean/max, and attention entropy mean/min.
\\

Value magnitude
&
Sink-position value-vector norm mean/max across attention heads.
\\

Attention--value
&
Sink attention-weighted value contribution mean/max and its ratio to the
total attention-output norm mean/max.
\\

Branch scales
&
Attention-output/input Frobenius-norm ratio, MLP-output/input ratio, and
block-update/input ratio.
\\

Compact hidden outlier
&
Hidden maximum absolute activation, hidden max/RMS ratio, and sink
hidden-token norm ratio.
\\

Hidden outlier + concentration
&
Compact hidden-outlier family plus hidden top-1 and top-5 elementwise
squared-energy ratios.
\\

Distributed multi-branch state
&
Max-absolute and max/RMS statistics for hidden input, attention output,
MLP output, block output, gate preactivation, up preactivation, and gated
MLP activation, together with sink-token norm ratios for hidden,
attention-output, MLP-output, and block-output branches.
\\

\bottomrule
\end{tabular}
\end{table}

The direct Qwen3 timing audit uses a separate stringent marker,

\begin{equation}
\frac{\max |H^{(\ell)}|}
{\operatorname{median}|H^{(\ell)}|}
\ge 10^4,
\end{equation}

and this marker should not be confused with the compact hidden-outlier family
used in the conditional analyses.

\subsection{Same-layer evaluation and implementation checks}
\label{app:same-layer-implementation}

All pre-sink analyses use prompt-grouped outer and inner splits,
fold-local preprocessing, and held-out prediction. For the model--layer
deconfounding sensitivity, let $c=(m,\ell)$ index a matched cell and let
$S_{icj}$ denote spectral feature $j$. Within each inner or outer training
split we compute

\begin{equation}
\mu^{\rm train}_{cj}
=
\frac{1}
{|\mathcal I_c^{\rm train}|}
\sum_{i\in\mathcal I_c^{\rm train}}
S_{icj},
\end{equation}

and replace validation/test features by

\begin{equation}
S_{icj}-\mu^{\rm train}_{cj}.
\end{equation}

The z-scored variant additionally divides by the corresponding training-cell
standard deviation. No held-out prompt contributes to these statistics.

Separately, cell-stratified evaluation computes AP inside each eligible
matched model--layer cell before macro or row-weighted aggregation. Because
model and layer are constant within such a cell, any cell-specific constant
cannot improve the within-cell ranking.

A group is marked \emph{non-estimable} if prompt-grouped nested cross-validation cannot keep both classes in every required split. We do not replace the primary procedure with a weaker validation rule simply to retain such groups. Attention, spectral, and activation tables are aligned by model,
prompt, and layer identifiers before fitting.

As an implementation audit, the original Qwen2.5 alignment check yields
maximum absolute discrepancies of $8.39\times10^{-8}$ for continuous
attention and $6.81\times10^{-8}$ for the sink score. Direct recomputation
for Qwen3 reproduces the stored attention, spectral $\alpha$,
spectral-$\alpha$ shift, and sink score exactly. The supplementary package
records prompt hashes, matched-cell counts, random seeds, input hashes, and
feature manifests.

\FloatBarrier


\subsection{Model-level sink patterns used for the main comparison}
\label{app:model-regimes}

The same-layer analysis requires both a measurable interval before sink formation and prompt-to-prompt variation in the formation layer. Table~\ref{tab:regimes} summarizes the eight model settings used to determine where that comparison is available.

\begin{table}[htbp]
\centering
\caption{Sink-formation patterns across the evaluated models. The first four provide the main same-layer comparison; the remaining models show settings where the same target is difficult to define or estimate.}
\label{tab:regimes}
\scriptsize
\setlength{\tabcolsep}{3.0pt}
\begin{tabular}{lccp{0.43\linewidth}l}
\toprule
Model & Prompts & $T$ & Observed sink behavior & Role \\
\midrule
Qwen2.5-7B & 50 & 64 & sinks often form near layer~4, with prompt variation & main \\
Qwen3-8B & 50 & 64 & sinks form mainly near layers~3 or~7 & main \\
GLM-4-32B & 50 & 64 & delayed formation with limited timing variation & main \\
GLM-Z1-32B & 50 & 64 & sink timing varies substantially across prompts & main \\
Gemma~3-12B & 20 & 64 & sinks appear early and strengthen later & boundary \\
Llama~3.1-8B & 20 & 64 & sinks are present at the first measured layer & boundary \\
GPT-OSS-20B & 20 & 256 & visibility of position~0 depends on attention type & boundary \\
Qwen3.6-27B & 50 & 64, 256 & sink-like states can appear, disappear, and reappear & boundary \\
\bottomrule
\end{tabular}
\end{table}

\FloatBarrier

\section{Additional Evidence for the Pre-Sink Spectral Change}
\label{app:warning-full}

The main text shows that the hidden-state spectrum contains information about later sink formation beyond current attention, with the clearest prompt-specific signal about two to three layers before the sink appears. This section reports the full horizon grid and tests whether that result depends on the attention baseline, one model, or residual differences between layers.

\subsection{Full horizon results}
\label{app:full-horizon-results}

Table~\ref{tab:primary-horizons} gives the complete primary matched-depth
horizon grid. The primary pooled confidence intervals exclude zero for both
Exact and Within targets through horizon~4 and include zero at horizon~6.

\begin{table}[t]
\centering
\caption{
\textbf{The pooled spectral gain is visible through horizon~4 and is absent
by horizon~6.}
Complete pooled matched-depth results for the pre-sink transition.
AP is reported for current attention ($\Afeat$) and current attention plus
hidden-state spectral geometry ($\Afeat+\Sfeat$), followed by paired
$\Delta$AP and prompt-bootstrap $95\%$ confidence intervals.
}
\label{tab:primary-horizons}
\scriptsize
\setlength{\tabcolsep}{4.0pt}
\begin{tabular}{llrrrr}
\toprule
Target & $h$ & $\Afeat$ & $\Afeat+\Sfeat$ & $\Delta$AP & $95\%$ CI \\
\midrule
Exact & 1 & .857 & .913 & +.056 & [.028,.087] \\
Exact & 2 & .826 & .867 & +.042 & [.011,.073] \\
Exact & 3 & .691 & .799 & +.108 & [.034,.162] \\
Exact & 4 & .754 & .824 & +.070 & [.010,.135] \\
Exact & 6 & .850 & .854 & +.005 & [$-.022$,.040] \\
\midrule
Within & 1 & .857 & .913 & +.056 & [.026,.088] \\
Within & 2 & .822 & .894 & +.073 & [.036,.113] \\
Within & 3 & .824 & .876 & +.052 & [.021,.085] \\
Within & 4 & .814 & .892 & +.078 & [.034,.122] \\
Within & 6 & .905 & .907 & +.002 & [$-.014$,.015] \\
\bottomrule
\end{tabular}
\end{table}

The pooled improvements remain visible through horizon~4 and disappear by horizon~6. The stricter within-group analyses below give the more conservative two--three-layer timing emphasized in the main text.

\FloatBarrier

\subsection{Expanded Qwen3 cohort and post-audit holdout}
\label{app:qwen3-expanded}

To increase prompt-level support without changing the $T=64$ measurement rule,
we apply the same Qwen3 pipeline to 500 frozen WikiText prompts that are
disjoint from the original 50-prompt cohort. Their onset counts are
$\{2:2,3:41,4:1,7:456\}$. The largest exact-horizon gain occurs at Exact~2,
where the full attention baseline gives $\Delta\mathrm{AP}=+.232$
$[.117,.340]$ and the continuous-attention-only baseline gives $+.195$
$[.062,.322]$. The corresponding full-baseline gains are $+.033$
$[-.038,.104]$ at Exact~1, $+.079$ $[-.046,.199]$ at Exact~3, and
approximately zero at Exact~4 and Exact~6. Thus the expanded cohort is
consistent with a short transition rather than a monotone long-horizon gain.

The 500-prompt set contains all 128 prompts used in an earlier strengthening
audit. To avoid treating that subset as independent evidence, we explicitly
partition it out using prompt hashes and rerun the frozen analysis on the
remaining 372 prompts. Their onset counts are $\{2:1,3:33,4:1,7:337\}$. The
full-baseline Exact~2 estimate remains positive at $+.102$ but its bootstrap
interval includes zero, $[-.035,.223]$; the continuous-attention-only value is
$+.020$ $[-.108,.132]$. Within~2 and Within~3 are also near zero on this
post-audit holdout. We therefore report the 500-prompt result as additional
support for the short-window pattern but not as a second independent
replication of the earlier 128-prompt audit, and we treat the size of the
predictive gain as cohort dependent.

\FloatBarrier

\subsection{Removing threshold-derived attention features}
\label{app:continuous-only}

The primary onset label is defined from a thresholded sink score, and the
attention family $\Afeat$ contains threshold-derived sink-score features. To
test whether this reuse creates a circular advantage for the spectral family,
we replay the historical primary pipeline after removing every current and
backward-difference feature of the form
\texttt{sink\_score\_eps\_*}. No replacement feature is added. Before this
sensitivity was run, the stored historical out-of-fold predictions were
reproduced to machine precision, fixing the exact primary preprocessing and
nested-CV pipeline.

\begin{table}[htbp]
\centering
\caption{
\textbf{The short-range spectral gain remains after threshold-derived
attention features are removed.}
The baseline retains continuous attention, enrichment, depth, fixed effects,
and the original nested-CV procedure.
}
\label{tab:continuous-only}
\scriptsize
\setlength{\tabcolsep}{4.0pt}
\begin{tabular}{lrrrr}
\toprule
Target & $\Afeat_{\rm cont}$ & $\Afeat_{\rm cont}+\Sfeat$ &
$\Delta$AP & $95\%$ CI \\
\midrule
Exact 1  & .833 & .918 & +.085 & [.051,.123] \\
Within 2 & .793 & .889 & +.096 & [.055,.137] \\
Within 3 & .787 & .867 & +.080 & [.042,.121] \\
Within 4 & .802 & .896 & +.094 & [.045,.141] \\
Within 6 & .899 & .912 & +.013 & [$-.003$,.031] \\
\bottomrule
\end{tabular}
\end{table}

The short-range spectral improvement remains after removing threshold-derived attention features. The Within~6 interval still includes zero. The result therefore does not arise from using the same thresholded quantity both to define sink formation and to describe current attention.

\FloatBarrier

\subsection{How much support comes from each model?}
\label{app:matched-support}

The amount of matched support varies substantially across models and
horizons. Table~\ref{tab:matched-support-app} makes this concentration
explicit. In particular, three models contribute only one matched layer to
the Exact~1 analysis, and GLM-4 contributes only three positive rows.

\begin{table}[t]
\centering
\caption{
Matched support by model and horizon. Entries in the four middle columns are
matched layers / positive rows. The final column reports the model-level
Exact~1 $\Delta$AP and $95\%$ confidence interval.
}
\label{tab:matched-support-app}
\small
\setlength{\tabcolsep}{3.1pt}
\begin{tabular}{lccccr}
\toprule
Model
& Exact 1
& Within 2
& Within 3
& Within 4
& Exact-1 $\Delta$AP [CI] \\
\midrule

GLM-4
& 1/3
& 1/3
& 1/3
& 1/3
& $+.203$ [$-.033,.951$] \\

GLM-Z1
& 5/43
& 7/84
& 8/118
& 9/151
& $+.032$ [$-.018,.097$] \\

Qwen2.5
& 1/36
& 2/72
& 3/108
& 4/144
& $+.007$ [$0,.020$] \\

Qwen3
& 1/10
& 2/20
& 3/30
& 3/30
& $-.094$ [$-.288,.025$] \\

\bottomrule
\end{tabular}
\end{table}

The model-level Exact~1 estimates should therefore not be interpreted as four
replicated standalone effects. The Qwen2.5 interval touches zero, the GLM-Z1
interval crosses zero, GLM-4 is extremely imprecise, and Qwen3 has a negative
point estimate with a wide interval. Our Exact~1 claim is therefore a pooled
statement. The better-supported temporal conclusion concerns the bounded
multi-block interval, for which matched support increases substantially at
the Within~2 and Within~3 horizons.

\FloatBarrier

\subsection{Removing residual model--layer differences}
\label{app:remove-residual-model-layer}

The primary estimator pools predictions across matched cells. To test whether
model--layer-specific spectral location contributes to this result,
Table~\ref{tab:cell-deconfounding} reports separately refit fold-local
centering and z-scoring analyses together with strict cell-stratified AP.
The untransformed estimates from this sensitivity pipeline are not used to
replace the primary values in Table~\ref{tab:primary-horizons}.

\begin{table}[htbp]
\centering
\caption{
\textbf{The strict prompt-specific signal is clearest at Within~2 and
Within~3 and is not evident at Within~4.}
Entries are $\Delta$AP with prompt-bootstrap $95\%$ confidence intervals.
``Centered'' and ``z-scored'' are pooled predictions after fold-local
model--layer transformation of spectral features. Cell-stratified columns
aggregate AP differences evaluated entirely within matched cells.
}
\label{tab:cell-deconfounding}
\scriptsize
\setlength{\tabcolsep}{2.7pt}
\begin{tabular}{lcccc}
\toprule
Target
& Centered pooled
& Z-scored pooled
& Cell macro
& Cell row-weighted \\
\midrule

Exact 1
& $+.057$ $[.017,.098]$
& $+.040$ $[.001,.078]$
& $+.169$ $[-.002,.217]$
& $+.096$ $[-.023,.151]$ \\

Within 2
& $+.068$ $[.029,.109]$
& $+.022$ $[-.018,.065]$
& $+.107$ $[.048,.155]$
& $+.102$ $[.048,.149]$ \\

Within 3
& $+.054$ $[.024,.088]$
& $+.018$ $[-.010,.046]$
& $+.053$ $[.002,.101]$
& $+.053$ $[.006,.105]$ \\

Within 4
& $+.068$ $[.034,.102]$
& $+.014$ $[-.028,.055]$
& $-.021$ $[-.037,.029]$
& $-.003$ $[-.016,.019]$ \\

\bottomrule
\end{tabular}
\end{table}

The Exact~1 cell-macro point estimate is particularly sensitive to very small
cells and is therefore not interpreted by its magnitude. More generally,
these controls support a hierarchy of claims rather than a single universal
horizon. Removing model--layer means preserves the pooled increment through
four blocks. Under the stricter requirement that ranking occur entirely
within matched cells, the cleanest prompt-specific evidence occurs at
Within~2 and Within~3, while Within~4 is no longer evident.

Cell-wise z-scoring further attenuates the longer horizons. Because this
transformation changes both model--layer-specific scale and the effective
regularization geometry, we do not interpret the attenuation as a unique
decomposition of stage and prompt information. We therefore use the
two--three-block interval as the cleanest estimate of the prompt-specific
transition window.

\FloatBarrier

\subsection{Alternative sink definitions and equal model weighting}
\label{app:sustained-onset}

First crossing is an operational definition of when the sink first forms, not an assumption that the state is permanently absorbing. We therefore define a
two-block sustained onset

\begin{equation}
\tau_i^{(2)}
=
\min\{\ell:
s^{(i,\ell)}_{\varepsilon}\ge\theta,\;
s^{(i,\ell+1)}_{\varepsilon}\ge\theta\},
\end{equation}

and analogously audit three consecutive blocks. Under the two-block
definition, the strict within-cell macro increments are

\begin{equation}
\begin{aligned}
&+.165\;[.031,.275],\quad +.217\;[.105,.258],\quad
 +.079\;[.032,.135],\\
&+.045\;[-.017,.108],\quad -.022\;[-.088,.029].
\end{aligned}
\end{equation}

for Exact~1, Within~2, Within~3, Within~4, and Within~6, respectively. Thus
the robust strict localization remains concentrated at approximately two to
three blocks. Requiring three consecutive sink-like blocks gives the same
qualitative localization and is treated as a secondary sensitivity.

The episode audit also clarifies what ``onset'' means across models. Qwen2.5 usually remains in the sink-like state after the first crossing, and Qwen3 does so for most prompts. GLM-4 usually exhibits a long first episode with brief
interruptions/re-entry, whereas GLM-Z1 is genuinely episodic: every audited
prompt has multiple sink-like episodes under the primary threshold. We therefore use \emph{first crossing} when discussing the primary target and reserve ``persistent onset'' for settings in which persistence is actually observed.

Unequal row counts across models are another possible pooled confound.
Reweighting models equally leaves the spectral increment positive with
bootstrap intervals above zero at the main horizons: $+.042$ $[.017,.070]$
for Exact~1, $+.056$ $[.028,.088]$ for Within~2, $+.037$ $[.014,.059]$ for
Within~3, and $+.061$ $[.030,.092]$ for Within~4. Model-level effects remain
heterogeneous, but the pooled transition does not depend on one model
contributing more rows.

\FloatBarrier

\subsection{Nonlinear attention baselines and a prompt-shuffling test}
\label{app:nonlinear-baselines}

\begin{figure}[htbp]
\centering
\includegraphics[width=0.72\linewidth]
{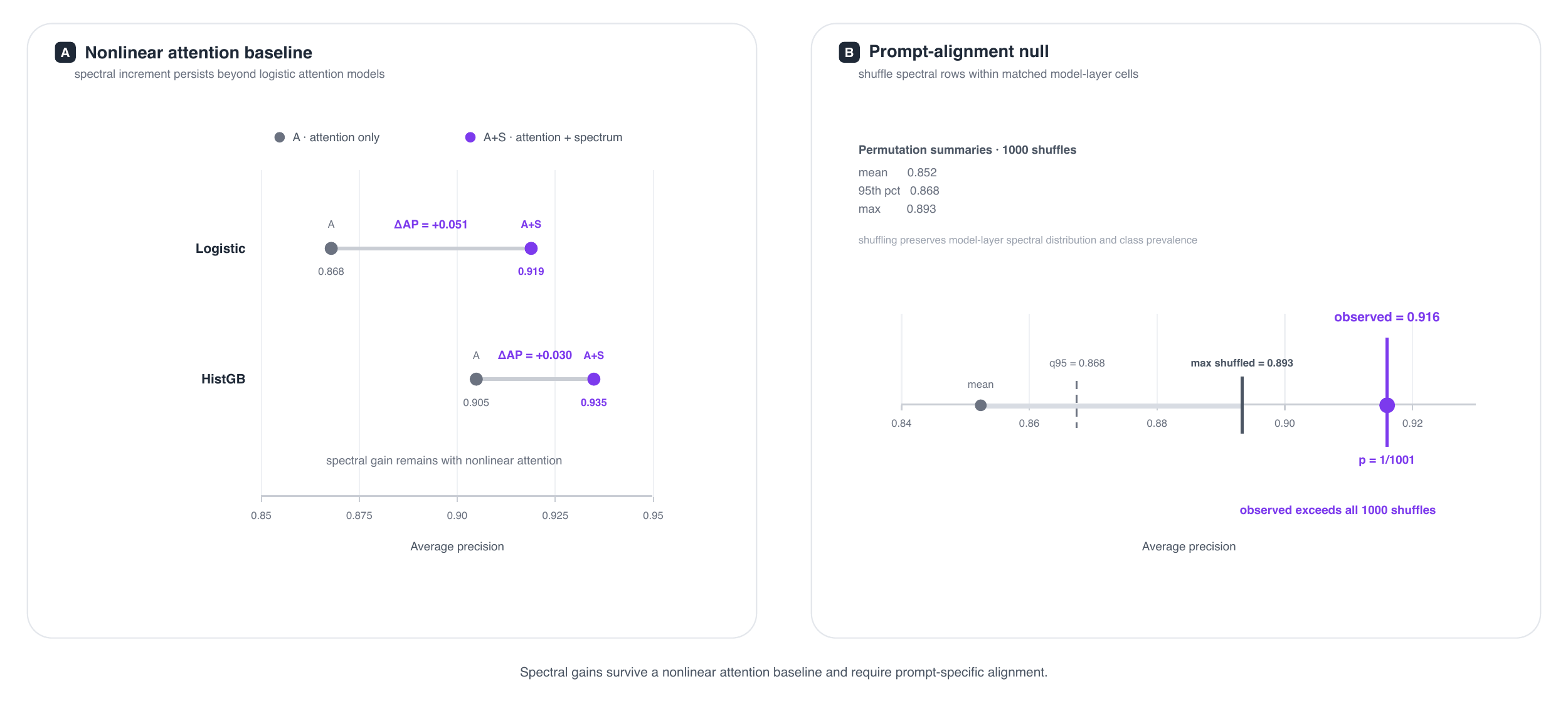}
\caption{
\textbf{Spectral gains survive a nonlinear attention baseline and depend on
prompt-specific alignment.}
\textbf{(A)} Adding spectral features improves average precision under both the
logistic attention baseline and a nonlinear HistGradientBoosting baseline,
showing that the spectral increment is not specific to the linear classifier.
\textbf{(B)} In the layer-preserving permutation test, complete spectral-feature
rows are shuffled across prompts within each matched model--layer cell,
preserving cell-level spectral distributions and class prevalence while
destroying prompt-specific alignment. The observed $A+S$ performance
($\mathrm{AP}=0.9160$) exceeds the maximum of all 1000 shuffled replicates
($0.8933$); the permutation mean is $0.8524$, the $95$th percentile is
$0.8675$, and the add-one Monte-Carlo value is $p=1/1001$.
}
\label{fig:app-warning-null}
\end{figure}

For each permutation replicate, the complete spectral-feature row is shuffled
across prompts separately within each model--layer cell. This preserves the
spectral distribution and class prevalence associated with every cell while
destroying prompt-specific spectral alignment.

The observed AP is $0.9160$, the permutation mean is $0.8524$, the 95th
percentile is $0.8675$, and the maximum among 1000 shuffled replicates is
$0.8933$. Thus the observed statistic exceeds every permutation replicate,
with add-one Monte-Carlo

\begin{equation}
p=\frac{1}{1001}.
\end{equation}

This permutation pipeline is separately refit, so its absolute AP should not
be substituted for the primary pooled value in
Table~\ref{tab:primary-horizons}.

\FloatBarrier

\subsection{Leave-one-model-out sensitivity}
\label{app:lomo-subsec}

Table~\ref{tab:lomo} reports complete leave-one-model-out refits. These use
a separately refit sensitivity pipeline and therefore test dependence on
model inclusion rather than re-estimating the primary pooled result.

\begin{table}[t]
\centering
\caption{
\textbf{No single model is required for the pooled Exact~1 effect.}
Leave-one-model-out $\Delta$AP with prompt-bootstrap $95\%$ confidence
intervals.
}
\label{tab:lomo}
\scriptsize
\setlength{\tabcolsep}{3.0pt}
\begin{tabular}{lcccc}
\toprule
Excluded model
& Exact 1
& Within 2
& Within 3
& Within 4 \\
\midrule

GLM-4
& \shortstack{$+.061$\\$[.028,.095]$}
& \shortstack{$+.130$\\$[.074,.187]$}
& \shortstack{$+.053$\\$[.022,.085]$}
& \shortstack{$+.049$\\$[-.008,.101]$} \\

GLM-Z1
& \shortstack{$+.155$\\$[.047,.262]$}
& \shortstack{$+.021$\\$[.006,.039]$}
& \shortstack{$+.038$\\$[.008,.067]$}
& \shortstack{$+.074$\\$[.035,.115]$} \\

Qwen2.5
& \shortstack{$+.165$\\$[.078,.264]$}
& \shortstack{$+.229$\\$[.138,.314]$}
& \shortstack{$+.166$\\$[.082,.241]$}
& \shortstack{$+.152$\\$[.074,.221]$} \\

Qwen3
& \shortstack{$+.030$\\$[.006,.056]$}
& \shortstack{$+.043$\\$[.018,.072]$}
& \shortstack{$+.057$\\$[.025,.092]$}
& \shortstack{$+.072$\\$[.031,.116]$} \\

\bottomrule
\end{tabular}
\end{table}

Every leave-one-model-out Exact~1 point estimate is positive. The
Within~4 interval crosses zero only when GLM-4 is excluded. These results
indicate that no single model drives the pooled effect; they do not imply
that every model exhibits a strong standalone effect at every horizon.

\FloatBarrier

\subsection{Generalization across prompt families}
\label{app:family-transfer}

A formal five-family GLM-Z1 analysis leaves out one complete prompt family in
each outer fold and selects regularization strength using only the remaining
training families. Table~\ref{tab:family-transfer} shows substantial
family- and horizon-dependence. The reasoning family has AP $=1$ for both
$\Afeat$ and $\Afeat+\Sfeat$ in these targets, so its
$\Delta$AP is exactly zero because of ceiling performance rather than failed
discrimination.

\begin{table}[t]
\centering
\caption{
\textbf{Prompt-family transfer is heterogeneous, with Within~3 the most
consistently positive horizon.}
GLM-Z1 five-family leave-one-family-out spectral increments. ``Positive''
counts held-out families with $\Delta$AP $>0$.
}
\label{tab:family-transfer}
\small
\setlength{\tabcolsep}{4.2pt}
\begin{tabular}{lrrrr}
\toprule
Target & Mean & Median & Range & Positive \\
\midrule

Exact 1
& +.150
& .000
& [$-.052,.760$]
& 2/5 \\

Within 2
& +.122
& +.009
& [$-.001,.547$]
& 3/5 \\

Within 3
& +.087
& +.040
& [$0,.316$]
& 4/5 \\

Within 4
& $-.005$
& +.019
& [$-.096,.027$]
& 3/5 \\

\bottomrule
\end{tabular}
\end{table}

The transfer signal is therefore heterogeneous across domains and horizons.
Within~3 is the most consistently positive target in this experiment, but we
do not interpret the result as a prompt-family-invariant property of the
transition.

\FloatBarrier


\section{Robustness to Sink Definition and Context Length}
\label{app:formation-robustness}

The main analysis treats onset depth as a continuous, prompt-dependent
quantity. This section examines whether the broad formation patterns used to
define the measurable pre-sink regime are stable to the operational sink
definition and to context length.

\subsection{Changing the sink threshold and query window}
\label{app:threshold-query-window}

The full 24-configuration grid varies

\begin{equation}
q_{\min}\in\{8,16\},
\qquad
\varepsilon\in\{0.2,0.3,0.4,0.5\},
\qquad
\theta\in\{0.4,0.5,0.6\}.
\end{equation}

Table~\ref{tab:threshold-stability} reports the resulting onset ranges.
Qwen2.5 retains median onset~4 in all 24 configurations and Qwen3 retains
median onset~7 in all 24 configurations, while GLM-4 and GLM-Z1 show wider
threshold-dependent changes in onset depth. Finer categorical descriptions
also vary within the grid.

\begin{table}[htbp]
\centering
\caption{
\textbf{Median onset is stable for Qwen2.5 and Qwen3 but more
threshold-dependent for the GLM models.}
Threshold and late-query sensitivity over 24 configurations. ``Label counts''
report operational regime assignments across the grid.
}
\label{tab:threshold-stability}
\scriptsize
\setlength{\tabcolsep}{3.0pt}
\begin{tabular}{lccc}
\toprule
Model
& Onset fraction range
& Median-onset range
& Label counts across 24 cells \\
\midrule

Qwen2.5
& [.72,1.00]
& 4--4
& fixed-depth partial 19; clustered 5 \\

Qwen3
& [1.00,1.00]
& 7--7
& deterministic 14; clustered 10 \\

GLM-4
& [1.00,1.00]
& 5--8
& deterministic 15; clustered 9 \\

GLM-Z1
& [.98,1.00]
& 6--11
& dispersed 15; deterministic 5; clustered 4 \\

\bottomrule
\end{tabular}
\end{table}

The stability of the median onset for Qwen2.5 and Qwen3 supports the broad
formation pattern used in the main analysis. The variation in categorical
labels reinforces our choice to treat continuous onset distributions as the
primary object and regime labels as operational summaries.

\FloatBarrier

\subsection{What happens at longer context lengths?}
\label{app:longer-context}

An initial Qwen3 nested-context audit uses the same 30 source texts truncated
to 64, 256, and 1024 tokens. Median onset remains block~7 at all three lengths
under both fixed and proportional late-query windows. At 64 tokens, 28/30
prompts onset at block~7 and two at block~3. At 256 and 1024 tokens, all
30 prompts onset at block~7. This suggested that longer context might collapse
the prompt-to-prompt onset variation required by the matched-depth task.

We therefore repeat the $T=256$ Qwen3 audit on a frozen cohort of 500
long-form WikiText source texts constructed by sequentially concatenating
non-empty WikiText-103 rows until at least 1400 Qwen3 tokens are reached.
The construction exactly reproduces the original 30 long-context source texts
before extending the cohort. All 500 prompts cross the sink criterion:
499/500 first cross at block~7 and one at block~3. Hence 99.8\% of prompts
share the same onset depth. The few nominally mixed layers are generated by
that single exceptional prompt, so prompt-grouped matched-depth discrimination
is effectively non-identifiable. We therefore do not collect a 500-prompt
Qwen3 spectral cohort at $T=256$.

Using the same 500 frozen source texts, GLM-Z1 remains identifiable. Its
two-block sustained-onset distribution is

\begin{equation}
\{6:100,\;7:260,\;11:66,\;12:3,\;16:71\},
\end{equation}

where the notation reports onset block and prompt count. The corresponding
long-context hidden-state experiment uses the same source cohort and evaluates
spectral information only after confirming mixed current-layer support.

\begin{table}[htbp]
\centering
\caption{
\textbf{At $T=256$, the strict GLM-Z1 signal again concentrates at
Within~2--3 and is near zero at longer horizons.}
Results use the frozen 500-source cohort. K1 is the first crossing; K2 requires
two consecutive sink-like blocks. Macro within-cell $\Delta$AP is reported
with prompt-bootstrap $95\%$ confidence intervals.
}
\label{tab:long-context-glmz1}
\scriptsize
\setlength{\tabcolsep}{3.8pt}
\begin{tabular}{lcccc}
\toprule
Onset definition & Within 2 & Within 3 & Within 4 & Within 6 \\
\midrule
K1
& $+.124$ $[.056,.241]$
& $+.066$ $[.038,.094]$
& $-.005$ $[-.026,.014]$
& $-.001$ $[-.012,.009]$ \\
K2
& $+.201$ $[.172,.244]$
& $+.079$ $[.063,.096]$
& $-.047$ $[-.097,.014]$
& $+.005$ $[-.005,.015]$ \\
\bottomrule
\end{tabular}
\end{table}

A Qwen2.5 timing-only audit on the same 500-source pool gives a similar
identifiability boundary. Of 500 prompts, 429 cross the primary sink criterion
and all 429 first cross at block~4; the remaining 71 do not cross in the
measured range. Because the crossing prompts have no onset-depth variation, we
do not collect a large spectral cohort for this setting.

The long-context result therefore separates two possibilities that are
otherwise easy to conflate. Qwen3 and Qwen2.5 do not provide evidence against a
spectral transition at $T=256$; their matched-depth questions are not estimable
because onset is nearly or completely depth locked. GLM-Z1 retains
prompt-dependent onset variation on the same source cohort, and in that
identifiable regime the strict spectral signal again localizes to approximately
two to three blocks.

\FloatBarrier


\section{Additional Tests of the Extreme-Activation Explanation}
\label{app:activation-state-redundancy}

The main text interprets the pre-sink spectral transition as a compact
observational readout of information broadly distributed across the current
activation state rather than as a detector of a single extreme activation.
This section reports two complementary tests of that interpretation. We first
ask how much incremental spectral information remains after conditioning on
other measurements of the current activation state. We then use Qwen3 to ask
whether a sharply defined extreme-activation event occurs early enough, and
varies enough across prompts, to account for prompt-specific sink timing.

\subsection{Conditioning on other activation measurements}
\label{app:conditioning-activations}

For a candidate activation-state family $\mathcal{M}$, define the conditional
spectral increment
\begin{equation}
\Delta_{\mathcal{M}}
=
\mathrm{AP}
(\Afeat+\mathcal{M}+\Sfeat)
-
\mathrm{AP}
(\Afeat+\mathcal{M}).
\label{eq:conditional-redundancy}
\end{equation}
A positive value means that spectral geometry retains predictive information
after conditioning on $\mathcal{M}$, whereas a value near zero means that
$\mathcal{M}$ already contains much of the predictive information contributed
by the spectral features. A negative value indicates no positive incremental
benefit from adding $\Sfeat$ in that fitted model. Thus,
$\Delta_{\mathcal{M}}$ measures predictive redundancy between the spectral
features and the conditioning state; it does not establish causal mediation.

\begin{figure}[htbp]
\centering
\includegraphics[width=0.96\linewidth]
{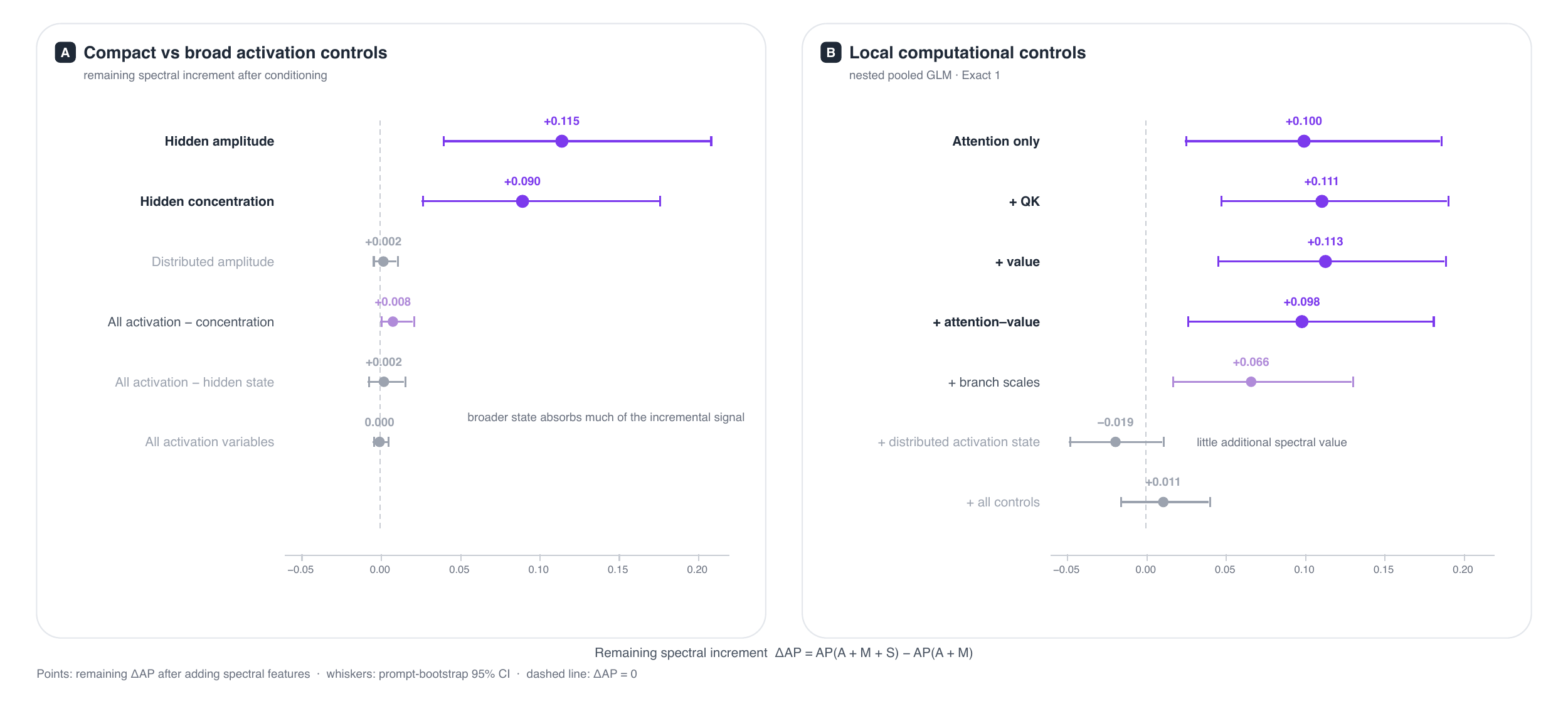}
\caption{
\textbf{Broader activation-state controls contain more of the predictive
information carried by the spectrum.}
\textbf{(A)} Compact hidden-state amplitude and concentration controls leave a
substantial positive spectral increment, whereas broader activation-state
descriptions reduce the remaining increment toward zero.
\textbf{(B)} In the nested pooled-GLM Exact~1 analysis, local computational
families---including query--key preference, value magnitude,
attention--value statistics, and branch scales---also leave a positive
spectral increment, while conditioning on the distributed activation state
leaves little additional spectral value.
Points show the remaining $\Delta$AP after adding spectral features,
horizontal bars show prompt-bootstrap $95\%$ confidence intervals, and the
vertical dashed line marks $\Delta$AP$=0$.
These comparisons measure predictive redundancy, not causal mediation.
}
\label{fig:activation-redundancy}
\end{figure}

Figure~\ref{fig:activation-redundancy} shows the same qualitative pattern in
two complementary conditioning analyses. Compact hidden-state controls leave a
substantial positive spectral increment. Likewise, conditioning on individual
local computational families---query--key preference, value magnitude,
attention--value statistics, or branch scales---does not consistently remove
the incremental spectral information. In contrast, broader descriptions of
the activation state reduce the remaining spectral increment to approximately
zero.

The result therefore suggests that the spectrum overlaps with information
distributed across the current activation state rather than merely tracking
one local activation or branch. This is a statement about predictive overlap,
not a causal decomposition: these conditioning analyses do not identify which
hidden-state variables generate the spectral change or mediate its relation to
sink formation.

\FloatBarrier

\subsection{Is the broader activation summary special?}
\label{app:broader-activation-special}

The distributed 18-dimensional activation-state control in
Figure~\ref{fig:activation-redundancy} removes most of the positive remaining
spectral increment. This overlap alone, however, does not show whether the
control is informative because of its specific content or simply because it is
a relatively rich feature family. We therefore evaluate three capacity-matched
control families on the same pooled GLM-4/GLM-Z1 cohort.

\paragraph{Prompt-aligned nuisance dimensions.}
We augment the compact hidden-state controls with 18 deterministic
prompt-aligned nuisance dimensions. Across 20 constructions, the remaining
spectral increments have means $+.160$, $+.177$, $+.163$, and $+.105$ for
Exact~1, Within~2, Within~3, and Within~4, respectively. None of the 20
nuisance controls reduces the remaining spectral increment as much as the real
structured 18-dimensional activation state at any horizon (add-one empirical
$p=1/21$). Thus, dimensionality and prompt alignment alone do not explain the
overlap between the broader activation measurements and the spectral features.

\paragraph{Alternative genuine activation summaries.}
We next draw 18-dimensional subsets from a 24-dimensional complementary
non-spectral activation pool. Across 25 draws, the structured state's
remaining spectral increment lies well inside the distribution obtained from
alternative activation summaries rather than forming a unique outlier. The
add-one fractions of random subsets that reduce the remaining spectral
increment at least as much as the structured state are .231, .154, .615, and
.769 across Exact~1 and Within~2--4.

\paragraph{Gaussian random projections.}
Finally, within each CV training fold we impute and scale the same
24-dimensional activation pool, project it to 18 dimensions with a Gaussian
random projection, and rescale before fitting. Across 25 projections, the
add-one fractions that reduce the remaining spectral increment at least as much
as the structured state are .154, .269, .462, and .846 for Exact~1 and
Within~2--4. The projected activation state can therefore recover much of the
same predictive content without preserving the semantics of the handcrafted
coordinates.

For completeness, we also shuffle complete structured 18-dimensional vectors
across prompts within each model--layer group. This destroys their
prompt-specific alignment while preserving the within-cell distribution of
the activation-state features. Across 200 shuffles, positive spectral
increments reappear at every horizon, and none of the shuffled controls reduces
the remaining spectral increment as much as the real prompt-aligned state
($p=1/201$). Thus, the reduction depends on genuine prompt-specific activation
information, whereas the nuisance, subset, and random-projection controls show
that this information is not unique to one handcrafted 18-dimensional basis.

Taken together, these controls support a simple interpretation. Broader,
prompt-specific descriptions of the activation state contain information that
overlaps substantially with the spectral features, but no single handcrafted
18-dimensional basis is uniquely responsible for that overlap. The spectrum
therefore appears to provide a compact readout of information distributed
across the current hidden state rather than a proxy for one specially chosen
activation family.

\FloatBarrier

\subsection{Qwen3: the extreme event is depth-locked across onset groups}
\label{app:qwen3-extreme-event}

Conditional redundancy provides one test of the single-outlier explanation.
Qwen3 provides a complementary temporal test: does the abrupt
extreme-activation event occur early enough, and vary enough across prompts,
to account for the prompt-specific difference in sink timing?

The direct Qwen3 audit recomputes exact hidden-activation statistics on the
same 50 formal prompts and reproduces the stored attention and spectral
records. Under the primary onset definition, 10 prompts form an attention sink
at block~3 and 40 at block~7.

At the $10^4$ maximum-to-median criterion, all 50 prompts remain below the
extreme-activation threshold before block~6 and first cross it at block~6.
We additionally sweep the criterion from $10^2$ to $3\times10^4$. For every
stringent threshold from $2\times10^3$ through $2\times10^4$, both onset
groups remain locked to first crossing at block~6. Lower thresholds detect
moderate pre-existing outliers earlier, while four prompts fail to cross the
$3\times10^4$ threshold.

\begin{figure}[htbp]
\centering
\includegraphics[width=0.96\linewidth]
{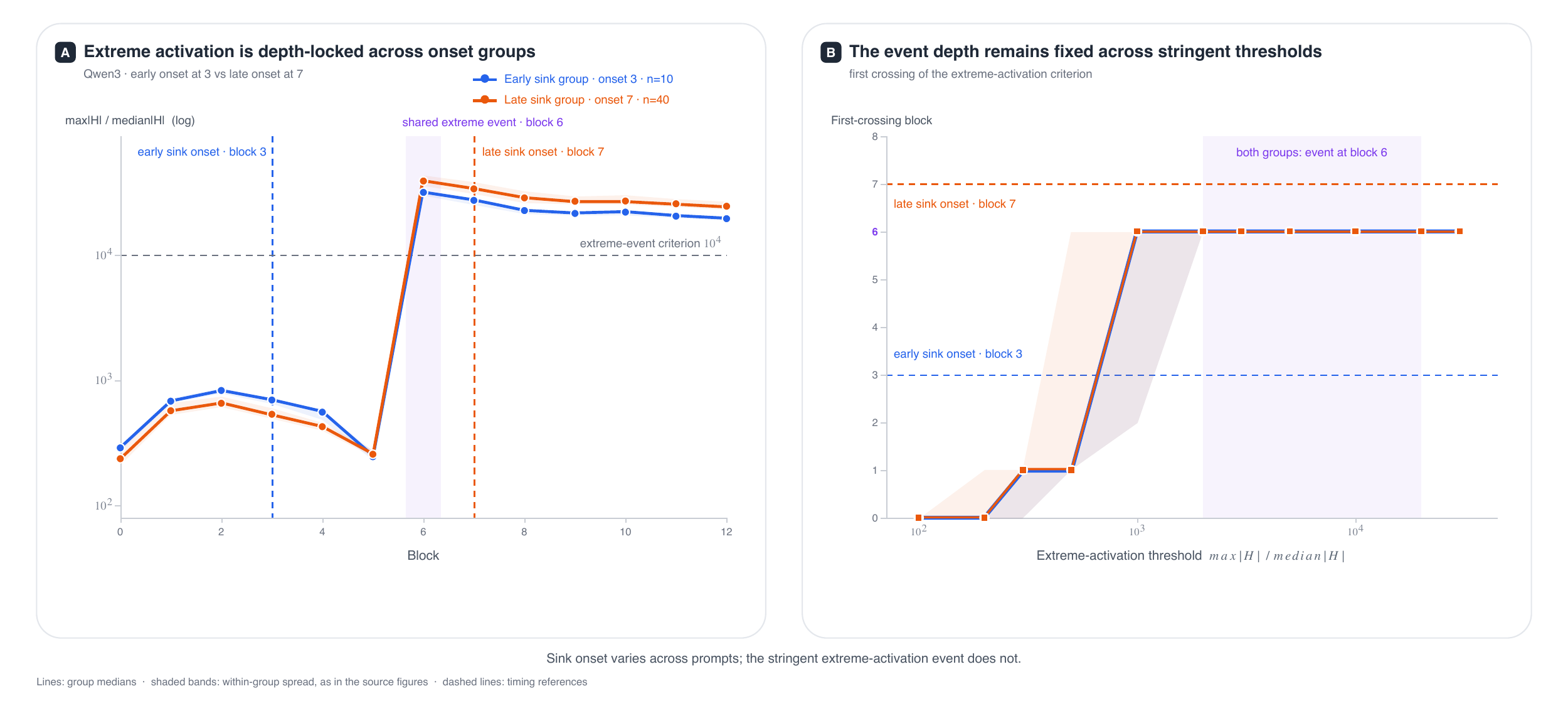}
\caption{
\textbf{Qwen3 sink timing varies across prompts while the stringent
extreme-activation event remains depth-locked.}
\textbf{(A)} Median maximum-to-median hidden-activation ratio, with
interquartile bands, for prompts whose sinks form at block~3 versus block~7.
Under the $10^4$ criterion, both groups exhibit the same abrupt
extreme-activation event at block~6, after the early-onset group has already
formed a sink and one block before the dominant late-onset group does.
\textbf{(B)} Sweeping the extreme-activation criterion shows that this shared
event depth is not specific to the single $10^4$ threshold. Across stringent
thresholds from $2\times10^3$ to $2\times10^4$, both onset groups first cross
at block~6 even though their sink onsets remain at blocks~3 and~7.
The figure therefore shows a timing mismatch: the stringent extreme-event
depth does not encode the prompt-specific difference in sink onset.
}
\label{fig:qwen3-extreme-event-timing}
\end{figure}

The timing mismatch rules out a simple explanation in which this stringent
extreme-activation event is a necessary precursor whose depth determines sink
onset for every prompt. In particular, the early-onset group has already formed
a sink three blocks before the event appears. For the dominant late-onset
group, however, the event remains closely associated with sink formation,
occurring one block beforehand. We therefore do not interpret this experiment
as showing that extreme activations are unrelated to sinks; rather, their
measured event depth cannot explain the prompt-to-prompt distinction between
block-3 and block-7 sink formation in this Qwen3 setting.

Together with the conditional activation-state analyses and the
capacity-matched controls, this timing result supports a consistent
interpretation: spectral geometry is a compact observational readout of
information broadly distributed across the current activation state, rather
than merely a detector of one abrupt extreme-activation event. These analyses
narrow the interpretation of the spectral precursor but do not identify a
unique causal mechanism for sink formation.


\section{Additional Intervention and Cross-Layer Results}
\label{app:causal-details}

This section provides the experiment-level evidence behind the causal summary
in the main text. We first give the exact intervention and matched-random
construction, then report the Qwen2.5 and GLM-Z1 tests, the frozen GPT-OSS
confirmation, the Qwen2.5 next-layer QK trace, direction-estimation stability,
and cross-layer propagation. Together these analyses separate the core causal
effect from the diagnostics used to explain its model- and layer-dependent
magnitude.

\subsection{Intervention construction}
\label{app:intervention-construction}

For every causal experiment we intervene on the centered post-block hidden
state $\widetilde H=U\operatorname{diag}(\sigma)V^\top$. The train-only
direction $d$ is constructed in normalized log-spectrum coordinates. For
delayed-onset models, $d$ is the difference between near- and far-state group
means within a fixed model--layer support cell. In the frozen Qwen2.5 layer-1
preparation, the near group is $1\le\tau-\ell\le3$, while the far group is
$\tau-\ell>3$ together with prompts that do not cross in the measured range;
already-crossed prompts are excluded. This rule is recovered exactly from the
stored preparation artifacts and reconstructs the frozen direction to machine
precision. For GPT-OSS, $d$ separates prompts with high and low later sink
strength for a fixed architecture-aware endpoint. The direction is mean
centered and normalized before use.

Only the leading spectral coordinates are edited. The token mean, left
singular subspace, and right singular subspace are preserved. Edited singular
values are rescaled to preserve centered spectral energy before
reconstruction. For each prompt, sign, and direction, a binary search chooses
the intervention coefficient $\eta$ so that the realized hidden perturbation

\begin{equation}
\frac{\|H'-H\|_F}{\|H\|_F}
\end{equation}

matches the requested budget to within $5\times10^{-4}$. Random spectral
directions are generated in the same spectral coordinate system and receive
the same realized-norm calibration. The zero-edit path is an identity bypass
rather than a reconstruction, avoiding numerical changes from an unnecessary
SVD round trip.

The hook is placed after block $\ell$. Therefore $A_\ell$ is computed before
the hidden state is edited, and any causal effect must enter through
$A_{\ell+1:}$ or later hidden-state computation. All reported causal runs
verify that the source-layer sink statistic remains unchanged.

\subsection{Qwen2.5 and GLM-Z1}
\label{app:qwen25-glmz1}

Qwen2.5 shows a reproducible directional effect under norm matching. The
observational near-state sign suppresses rather than increases later sink strength relative to the opposite sign, illustrating why the direction is
best treated as a state coordinate rather than a ``toward-sink'' vector.
The tie-aware random-direction empirical $p$ values are .0615, .0308, and
.0154 for $2.5\%$, $5\%$, and $10\%$ budgets, respectively, with natural
future directional gaps $-.00536$, $-.00833$, and $-.01726$.

GLM-Z1 provides an independent cross-model corroboration. Table~\ref{tab:glm-causal}
reports the results against 64 realized-norm-matched random directions.

\begin{table}[htbp]
\centering
\caption{
\textbf{GLM-Z1 separates from the matched random-direction null across the
tested budgets, with the smallest budget giving the cleanest functional-damage
profile.}
Natural gap is near-state minus far-state later sink strength. Random intervals
are the empirical $2.5$--$97.5$ percentile ranges over 64
realized-norm-matched spectral directions. The reported $p$ values use the
stored matched random-direction empirical test.
}
\label{tab:glm-causal}
\small
\begin{tabular}{lrrr}
\toprule
Budget & Natural gap & Random 95\% interval & Empirical $p$ \\
\midrule
.025 & $-.01525$ & $[-.01205,.00946]$ & .0308 \\
.050 & $-.02976$ & $[-.02773,.02174]$ & .0308 \\
.100 & $-.05022$ & $[-.04574,.03433]$ & .0462 \\
\bottomrule
\end{tabular}
\end{table}

The combined through-origin dose slope is negative and lies outside the
paired 64-direction random null (empirical $p=.0308$). A post-hoc distance-to-onset
audit shows, however, that the pooled effect is strongly state dependent. For
prompts with $r=\tau-\ell\le4$, the natural gaps are small and positive across
the three budgets; for $r\ge8$, they are $-.0531$, $-.0990$, and $-.1750$,
each outside the corresponding matched random-direction range. These subgroup
results are used as a heterogeneity diagnostic rather than as separately
confirmed effects. We therefore do not claim that maximal causal
susceptibility universally coincides with the two--three-block observational
precursor window.

\subsection{GPT-OSS frozen confirmation}
\label{app:gptoss-frozen}

GPT-OSS alternates sliding and full attention. The preparation stage therefore
scores each candidate source layer by mean sink strength over its next three full-attention layers. The selected sources are layer~9 (full attention) and
layer~10 (sliding attention), both with future targets $11,13,15$; their
train-derived spectral directions have cosine $0.9907$. A consensus direction
is frozen before intervention.

The initial six-prompt pilot finds a much larger natural directional effect at
layer~10 than at layer~9. A layer-8/layer-9/layer-10 sandwich extension shows
that the result is not a generic sliding-versus-full distinction: the
layer-8 sliding source is approximately null, the layer-9 full source is
small, and the layer-10 sliding source immediately before full-attention
layer~11 is large.

All design choices are then frozen and evaluated on a disjoint 30-prompt
cohort from a separate frozen WikiText prompt pool using 64 random directions. Table~\ref{tab:gptoss-fresh-causal}
summarizes the confirmatory result.

\begin{table}[htbp]
\centering
\caption{
\textbf{The frozen GPT-OSS intervention gives a positive fresh-cohort effect
on all 30 prompts.}
The test is run at layer~10 with $\rho=.025$ and uses mean sink strength over
full-attention layers 11, 13, and 15 as the endpoint.
}
\label{tab:gptoss-fresh-causal}
\small
\begin{tabular}{lr}
\toprule
Quantity & Value \\
\midrule
Prompts & 30 \\
Natural high-minus-low effect & $+.05625$ \\
Bootstrap 95\% CI & $[.04983,.06267]$ \\
Prompt sign consistency & 30 positive / 0 zero / 0 negative \\
Random median & $-.00269$ \\
Random 2.5--97.5 percentile range & $[-.05538,.05548]$ \\
One-sided empirical $p$ & .0308 \\
Two-sided empirical $p$ & .0462 \\
Mean natural KL & .0107 \\
Maximum source-layer sink change & 0 \\
\bottomrule
\end{tabular}
\end{table}

The natural KL is comparable to the random-direction distribution, so the
pre-specified directional effect is not explained by a systematically larger
generic output disturbance. The positive sign is also consistent across every
fresh prompt.

\subsection{Local Qwen2.5 query--key trace}
\label{app:qwen25-qk}

To identify the first attention computation changed by the Qwen2.5 layer-1
intervention, we compare $+d$, the unedited state, and $-d$ at layer~2 while
reusing the frozen direction and $\rho=.05$. The source-layer sink statistic is
unchanged exactly. Relative to $-d$, the $+d$ edit reduces mean position-0
attention at layer~2 by $-.02142$ with prompt-bootstrap $95\%$ CI
$[-.02419,-.01846]$; 48/50 prompts have a negative change. The corresponding
threshold-based sink-score difference is $-.02714$ $[-.04143,-.01357]$.

For each prompt and head, define the relative position-0 query--key preference
from the attention probabilities as
\begin{equation}
M_{qk}=\log p(k=0)-\log\max_{k\ne0}p(k),
\end{equation}
which equals the corresponding pre-softmax logit difference within a fixed
query row. Across prompt--head pairs, $\Delta M_{qk}$ tracks the change in
position-0 attention with correlation $r=.544$ and bootstrap $95\%$ CI
$[.395,.671]$; the two changes agree in sign for $97.7\%$ of pairs. The
position-0 value-vector norm does not show comparable directional agreement
(sign agreement $33.9\%$). We therefore use the query--key result as a local
computational trace of the Qwen2.5 suppression, not as a universal causal
decomposition of sink formation.

\subsection{Direction-estimation robustness}
\label{app:direction-stability}

For Qwen2.5, exact reconstruction of the stored layer-1 preparation yields
cosine $0.9999999999999999$ with the frozen direction and maximum coordinate
error $1.29\times10^{-15}$. The full training support contains 25 near prompts
and 10 far prompts. Across 2000 repeated partitions into two non-overlapping
halves, the cosine between a half-data direction and the full-data direction
has median $.9947$ and empirical $2.5$--$97.5$ percentile interval
$[.9750,.9987]$. The stricter cosine between two independently estimated,
disjoint half-data directions has median $.9789$ and interval $[.9022,.9947]$;
the minimum across all 2000 partitions is $.808$. When the half-A direction is
projected onto the held-out half-B prompts, the near-minus-far projection gap
has median $.597$ and is positive in 2000/2000 partitions.

For GPT-OSS, 12 deterministic half-training directions are estimated from only
17 of the 35 preparation prompts and then evaluated with the frozen layer-10,
$\rho=.025$, 30-prompt test cohort, and layer-11/13/15 endpoint. Their cosine
with the full-data direction has median $.9858$ and empirical range
$[.9740,.9916]$. The median causal effect is $+.05634$ with range
$[+.05553,+.05660]$, compared with $+.05625$ for the full-data direction.
The effect-retention ratio has median $1.002$ and range $[.987,1.006]$, and all
12 directions preserve the positive causal sign. These are empirical ranges
over deterministic data splits, not confidence intervals; significance against
random directions is still provided by the separate 64-direction frozen test.

\subsection{How the perturbation changes across layers}
\label{app:perturbation-transport}

The causal effect of a spectral edit depends strongly on source layer. To
understand why, we compare propagated upstream perturbations with the direct
local edit at the same downstream hidden state.

In GPT-OSS, the layer-10 direct edit is the reference perturbation immediately
before full-attention layer~11. A layer-9 perturbation reaches $H_{10}$ with
relative hidden norm about $3.0\%$, and a layer-8 perturbation reaches it with
about $3.5\%$, both at least as large as the $2.5\%$ direct edit. The
projection of the propagated log-spectrum displacement onto the frozen
spectral direction retains approximately $100\%$ and $96\%$ of the direct
value, respectively. Nevertheless, the centered hidden displacement has
cosine only about $0.81$ (layer~9 source) and $0.65$ (layer~8 source) with
the direct layer-10 edit. The causal effect on layer~11 correspondingly
collapses or changes sign. This is consistent with block-wise reorientation of
the perturbation even when the targeted spectral coordinate is preserved.

Qwen2.5 shows a different form of transport. The direct layer-1 edit has
relative hidden norm about $2.51\%$. Injecting the same frozen spectral
direction one block earlier at layer~0 produces a layer-1 displacement of
about $2.84\%$, but only about $24\%$ of the direct spectral projection
survives and the cosine with the direct layer-1 hidden displacement is about
$0.18$. Here the intervening block both filters the targeted spectral mode and
strongly reorients the surviving perturbation.

These transport results do not constitute a separate random-direction
significance test. Their role is mechanistic: after causal specificity has
been established, they explain why the same nominal spectral deformation can
have sharply different downstream effects depending on where it is inserted.

\FloatBarrier


\section{Architecture-Specific Cases}
\label{app:architecture-boundaries}

The following models are included to delimit the conditions under which the
pre-sink transition and a persistent onset can be meaningfully defined. They
are not treated as additional independent phenomena or as members of the
four-model matched-depth cohort used to establish the primary transition.

Gemma forms sinks early and strengthens them later, Llama is already saturated at the first measured layer, GPT-OSS has a structural visibility constraint, and Qwen3.6 repeatedly enters and leaves sink-like states.

\subsection{Why GPT-OSS needs a different sink measure}
\label{app:gptoss-different-sink}

In GPT-OSS, attention architecture itself constrains whether the candidate
sink position can receive nonzero mass. Once position~0 leaves a local
sliding-attention window, zero attention to that position is structurally
enforced, whereas full-attention layers can continue to assign attention to it.
Consequently, absence of a measured sink at a local layer need not represent
the same latent state as below-threshold sink strength at a full-attention layer.

For this reason GPT-OSS is not included in the ordinary persistent-onset
matched-depth cohort. The causal experiments instead define an
architecture-aware endpoint using subsequent full-attention layers. This
turns the structural boundary into a controlled test: a spectral deformation
inserted after a local layer can be evaluated by whether it changes later
sink strength at globally visible full-attention layers. The resulting layer-10 intervention and
fresh-cohort confirmation are reported in
Appendix~\ref{app:causal-details}.

\begin{figure}[htbp]
\centering
\includegraphics[width=0.72\linewidth]
{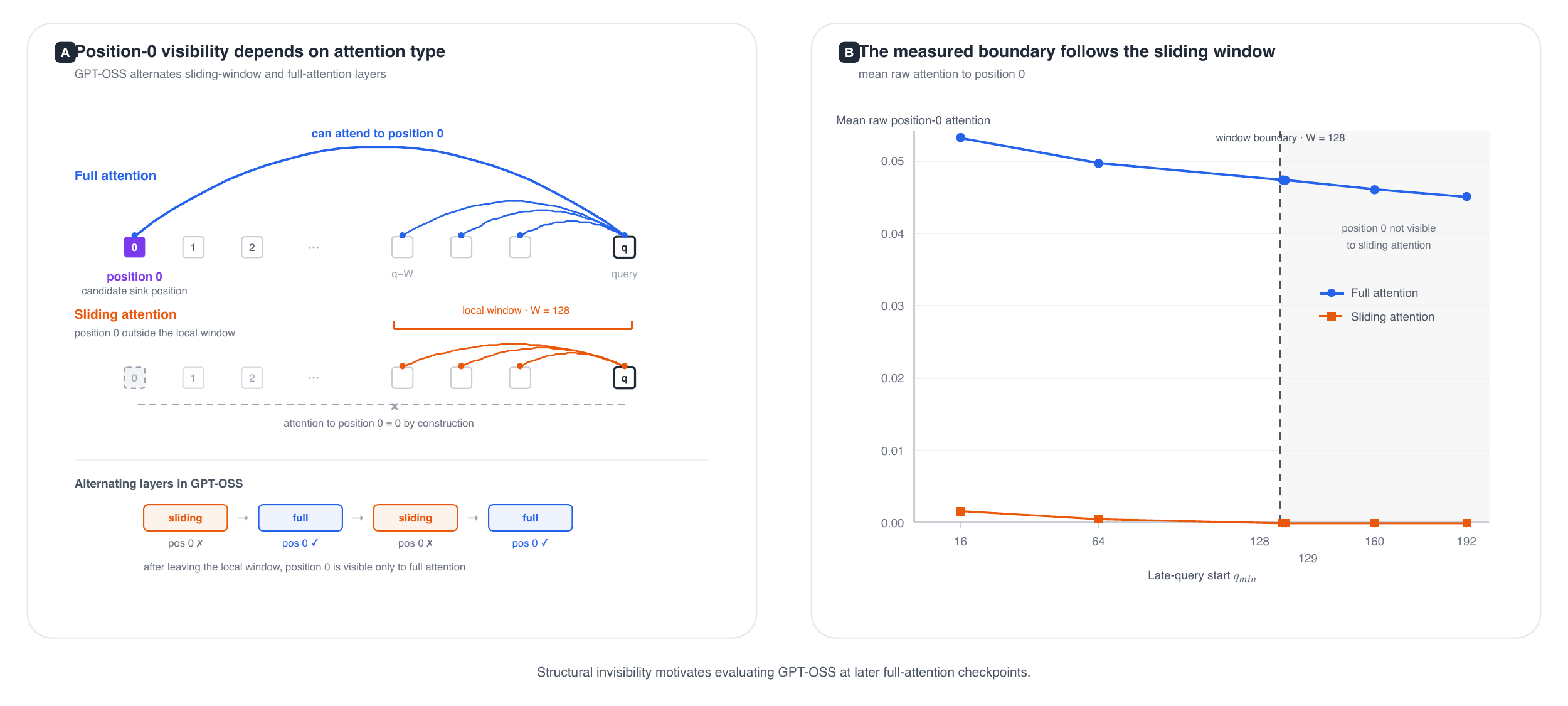}
\caption{
\textbf{GPT-OSS requires an architecture-aware sink measure because
position-0 visibility depends on attention type.}
\textbf{(A)} Full-attention layers can attend to position~0 from arbitrarily
late queries, whereas in a sliding-attention layer position~0 becomes
structurally inaccessible once it leaves the local window; zero attention in
that regime is therefore enforced by the architecture rather than evidence
that a sink-like state has disappeared.
\textbf{(B)} The measured position-0 attention follows this visibility
boundary: full-attention layers retain nonzero mass, while sliding-attention
layers approach structural zero once position~0 is outside the local window.
This motivates evaluating the GPT-OSS causal intervention at subsequent
full-attention checkpoints rather than defining a single persistent
position-0 sink onset across all layers.
}
\label{fig:app-gptoss}
\end{figure}

\FloatBarrier

\subsection{Why Qwen3.6 does not have one persistent sink-formation layer}
\label{app:qwen36-persistent-sink}

Qwen3.6-27B contains 64 raw blocks, of which 16 are full-attention
checkpoints at raw layers

\begin{equation}
3,7,\ldots,63.
\end{equation}

Under the primary $q_{\min}=16$ protocol at $T=64$, 45/50 prompts cross the
sink threshold at least once and 41 first cross at raw layer~15. At
$T=256$, only 27/50 prompts cross, with first crossings distributed across
raw layers 15, 47, 55, and 59 in counts 11/6/5/5.

The more stable qualitative property is not the depth of the first crossing
but its non-persistence. At $T=256$, accepted fractions rise to 11/50 at
layer~15 and fall to 0/50 at layer~19, then rise again to 11/50 at
layer~47 and fall to 0/50 at layer~51. Only 5/27 first crossings remain
accepted at the next full-attention checkpoint.

The later-query $q_{\min}=64$ analysis uses the 45 prompts with a nonempty
late-query region. It shifts first crossing later---2 prompts at layer~47,
5 at layer~55, and 10 at layer~59---while preserving the recurrent,
non-persistent pattern. Only 5/17 first crossings remain accepted at the
next full-attention checkpoint.

Thus exact first-crossing depth is sensitive to context length and query
window, whereas repeated appearance and disappearance at full-attention checkpoints is the more stable qualitative pattern.

\begin{figure}[htbp]
\centering
\includegraphics[width=0.80\linewidth]
{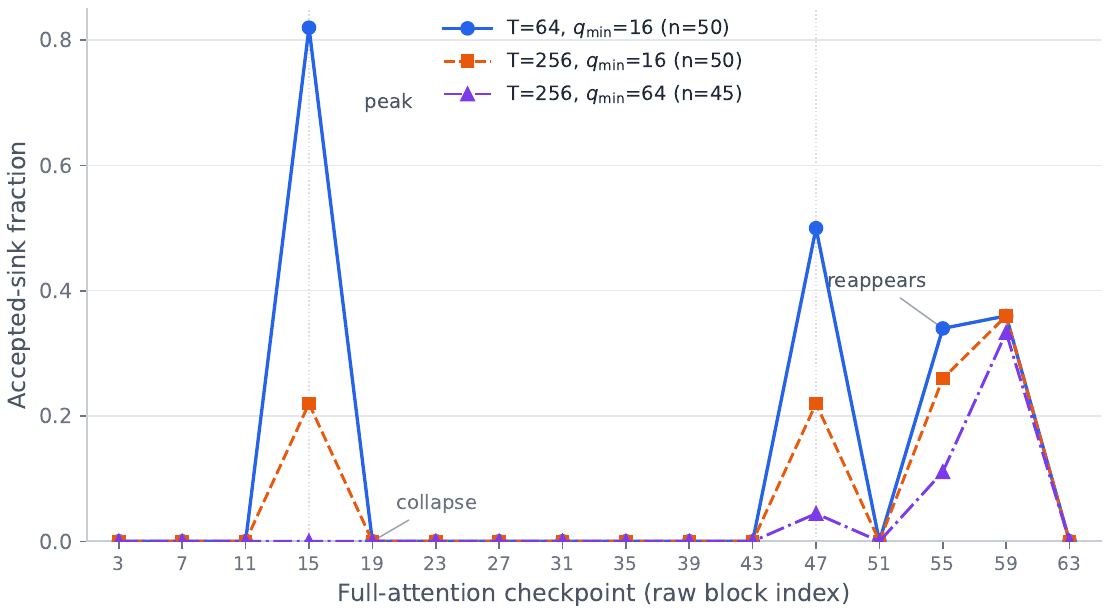}
\caption{
\textbf{Qwen3.6 hybrid-attention boundary.}
Accepted sink-state fractions are shown only at the 16 full-attention
checkpoints. The recurrent peak--collapse--reappearance pattern persists
across context length and query-window choices, whereas the location and
prevalence of the first threshold crossing vary substantially. This behavior
supports treating Qwen3.6 as a recurrent, architecture-gated sink regime
rather than an ordinary persistent-onset process.
}
\label{fig:app-qwen36-recurrence}
\end{figure}

\begin{table}[htbp]
\centering
\caption{
\textbf{Qwen3.6 first crossings are often non-persistent and shift with
context length and query window.}
``Next-checkpoint persistence'' is evaluated only for prompts with a first
crossing and an observed subsequent full-attention checkpoint.
}
\label{tab:qwen36-recurrence}
\scriptsize
\setlength{\tabcolsep}{3.4pt}
\begin{tabular}{lcccc}
\toprule
Setting
& Eligible prompts
& Any crossing
& First-crossing raw layers (counts)
& Next-checkpoint persistence \\
\midrule

$T=64,\ q_{\min}=16$
& 50
& 45
& 15 (41), 47 (2), 55 (1), 59 (1)
& --- \\

$T=256,\ q_{\min}=16$
& 50
& 27
& 15 (11), 47 (6), 55 (5), 59 (5)
& 5/27 \\

$T=256,\ q_{\min}=64$
& 45
& 17
& 47 (2), 55 (5), 59 (10)
& 5/17 \\

\bottomrule
\end{tabular}
\end{table}

\paragraph{No robust extension of the matched-depth signal.}

A fixed-target nested-CV analysis does not show a robust positive spectral
increment in the recurrent Qwen3.6 regime. At
$T=256,q_{\min}=16$, predicting the dominant layer-15 crossing from one,
two, or three raw blocks earlier gives

\begin{equation}
\Delta\mathrm{AP}
=
-.180\;[-.437,.037],
\qquad
+.179\;[-.066,.406],
\qquad
-.078\;[-.290,.149],
\end{equation}

respectively.

At $q_{\min}=64$, where the dominant crossing shifts to layer~59, the
corresponding increments are

\begin{equation}
-.006\;[-.081,.068],
\qquad
-.032\;[-.177,.078],
\qquad
-.136\;[-.272,.002].
\end{equation}

On the 45-prompt common cohort, a three-block $q_{\min}=16$ analysis yields
an isolated

\begin{equation}
+.230\;[.009,.563]
\end{equation}

with only six positives, but this does not replicate under $q_{\min}=64$.

We therefore do not interpret the current Qwen3.6 results as evidence for a
robust architecture-invariant spectral increment. Qwen3.6 does not provide
evidence against the pre-sink transition identified in the delayed
persistent-onset models; rather, it shows that the target being
measured---a single persistent sink onset---is itself not architecture
invariant.

\FloatBarrier


\section{Exploratory Jacobian Analysis}
\label{app:routes}

This exploratory analysis asks whether local Jacobian measurements explain the spectral change. It is not used as evidence for the main claim.

The five-route Jacobian variables alone predict spectral-$\alpha$ shifts with

\begin{equation}
R^2_{\rm OOF}=0.665,
\end{equation}

and the observed association exceeds a 1000-replicate within-prompt
circular-shift null, with

\begin{equation}
p=\frac{1}{1001}.
\end{equation}

However, a flexible model-specific polynomial-depth baseline achieves

\begin{equation}
R^2_{\rm OOF}=0.846,
\end{equation}

and combining depth with the route variables increases this only to

\begin{equation}
R^2_{\rm OOF}=0.850.
\end{equation}

Thus the route variables track spectral evolution, but provide only marginal
additional explanatory power beyond flexible model-specific depth dynamics.

\begin{figure}[htbp]
\centering
\includegraphics[width=0.72\linewidth]
{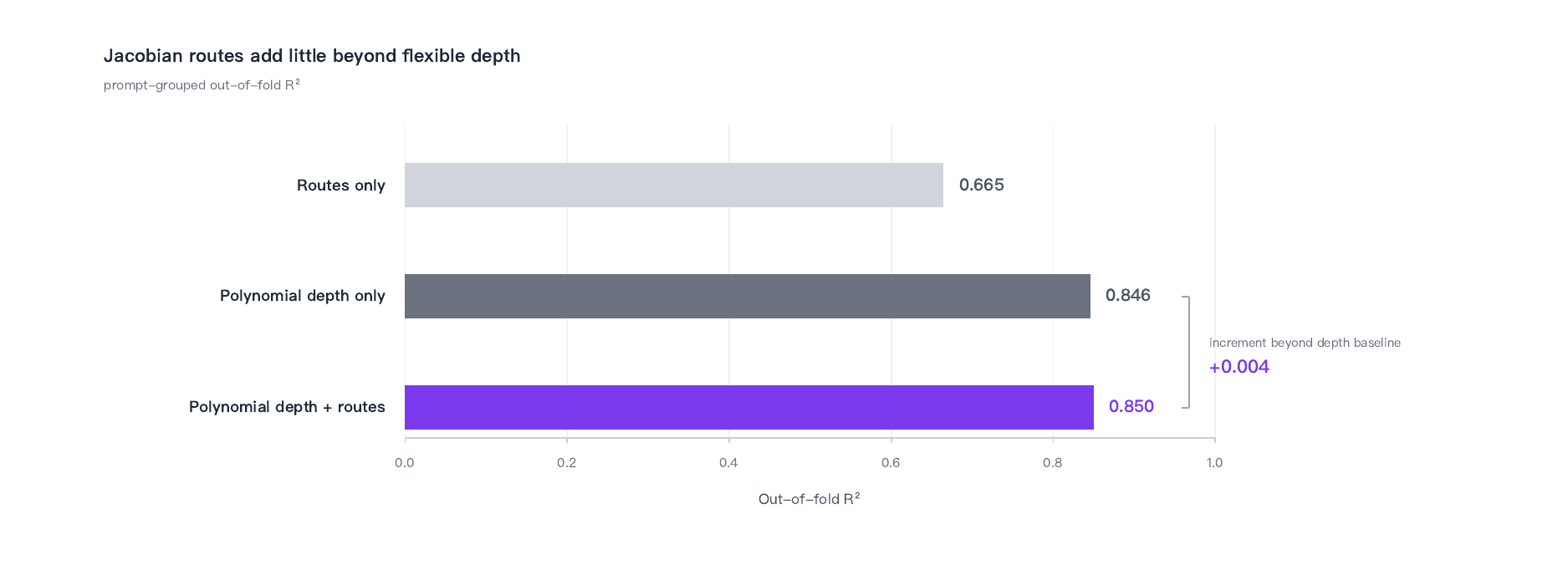}
\caption{
\textbf{Route-to-spectrum regression with nonlinear depth control.}
Jacobian route variables explain spectral-$\alpha$ shifts under
prompt-grouped out-of-fold evaluation, but most of this explanatory power is
shared with flexible model-specific nonlinear depth. Adding route variables
to the polynomial-depth baseline yields only a marginal improvement
($R^2_{\mathrm{OOF}}: 0.846 \rightarrow 0.850$; $\Delta R^2 = 0.004$).
}
\label{fig:app-routes}
\end{figure}

The Jacobian measurements therefore track the same broad evolution with depth but add little once flexible depth dynamics are known. We do not use this analysis as independent evidence for the mechanism behind sink formation.

\FloatBarrier


\section{Summary of Claims and Evidence}
\label{app:scope}

The main text reports only the anchor results needed to follow the argument.
Table~\ref{tab:complete-results} provides a single quantitative index of the
main observational, robustness, causal, mechanistic, and boundary experiments.
Unless otherwise stated, bracketed intervals are prompt-bootstrap 95\%
confidence intervals; empirical ranges are labeled explicitly.

\begingroup
\setlength{\tabcolsep}{2.5pt}
\begin{longtable}{
p{0.16\linewidth}
p{0.24\linewidth}
p{0.38\linewidth}
p{0.16\linewidth}}
\caption{\textbf{Complete quantitative summary of the main empirical evidence.}
The table consolidates the results used to support or delimit the paper's
claims; experiment-specific tables and figures provide the full context.}
\label{tab:complete-results}\\
\toprule
Analysis & Comparison / setting & Main quantitative result & Role in the paper \\
\midrule
\endfirsthead

\multicolumn{4}{l}{\small\textit{Table~\ref{tab:complete-results} continued}}\\
\toprule
Analysis & Comparison / setting & Main quantitative result & Role in the paper \\
\midrule
\endhead

\midrule
\multicolumn{4}{r}{\small Continued on next page}\\
\endfoot

\bottomrule
\endlastfoot

\multicolumn{4}{l}{\textbf{Observational evidence}}\\[2pt]

Primary pooled horizon grid
&
Current attention $\Afeat$ vs.\ attention plus spectrum
$\Afeat+\Sfeat$, matched by model and current layer
&
Exact~1--6 $\Delta$AP:
$+.056$, $+.042$, $+.108$, $+.070$, $+.005$.
Within~1--6:
$+.056$, $+.073$, $+.052$, $+.078$, $+.002$.
The Exact/Within intervals exclude zero through horizon~4 and include zero at
horizon~6.
&
Shows a short-range spectral increment beyond current attention.
\\

Strict within-cell timing
&
AP is computed only among prompts in the same model and current layer
&
Exact~1:
$+.169$ $[-.002,.217]$;
Within~2:
$+.107$ $[.048,.155]$;
Within~3:
$+.053$ $[.002,.101]$;
Within~4:
$-.021$ $[-.037,.029]$.
&
Localizes the clearest prompt-specific signal to roughly two to three layers.
\\

Continuous-attention control
&
Remove all threshold-derived sink-score features from the attention baseline
&
Exact~1:
$.833\rightarrow.918$ ($+.085$ $[.051,.123]$);
Within~2:
$.793\rightarrow.889$ ($+.096$ $[.055,.137]$);
Within~3:
$+.080$ $[.042,.121]$;
Within~4:
$+.094$ $[.045,.141]$;
Within~6:
$+.013$ $[-.003,.031]$.
&
Rules out reuse of the onset threshold as the explanation for the short-range
gain.
\\

Expanded Qwen3 cohort
&
Frozen $T=64$ analysis on 500 prompts, followed by an unseen 372-prompt
post-audit subset
&
$N=500$, Exact~2:
$+.232$ $[.117,.340]$;
continuous-attention-only:
$+.195$ $[.062,.322]$.
Unseen $N=372$, Exact~2:
$+.102$ $[-.035,.223]$;
continuous-only:
$+.020$ $[-.108,.132]$.
&
Shows a strong short-window result in the expanded cohort but weaker
out-of-audit predictive strength.
\\

Sustained-onset sensitivity
&
Require two consecutive sink-like layers instead of first crossing
&
Strict within-cell:
Exact~1 $+.165$ $[.031,.275]$;
Within~2 $+.217$ $[.105,.258]$;
Within~3 $+.079$ $[.032,.135]$;
Within~4 $+.045$ $[-.017,.108]$;
Within~6 $-.022$ $[-.088,.029]$.
&
Shows that the two--three-layer localization is not specific to first crossing.
\\

Nonlinear baseline
&
Logistic and HistGradientBoosting attention-only baselines, each compared with
the same model after adding spectrum
&
Logistic:
AP $.868\rightarrow.919$ ($\Delta$AP $=+.051$).
HistGB:
$.905\rightarrow.935$ ($\Delta$AP $=+.030$).
&
Shows that the spectral increment is not an artifact of a linear
attention-only classifier.
\\

Prompt-alignment null
&
Shuffle complete spectral rows across prompts within each matched model--layer
cell
&
Observed AP $=.9160$;
null mean $=.8524$;
95th percentile $=.8675$;
maximum $=.8933$;
add-one Monte-Carlo $p=1/1001$.
&
Shows that the gain depends on prompt-specific spectral alignment.
\\

Long-context GLM-Z1
&
Strict matched-layer analysis at $T=256$ on the frozen 500-source cohort
&
K1:
Within~2 $+.124$ $[.056,.241]$,
Within~3 $+.066$ $[.038,.094]$,
Within~4 $-.005$ $[-.026,.014]$,
Within~6 $-.001$ $[-.012,.009]$.
K2:
$+.201$ $[.172,.244]$,
$+.079$ $[.063,.096]$,
$-.047$ $[-.097,.014]$,
$+.005$ $[-.005,.015]$.
&
Shows the same short-window localization when long-context onset variation
remains identifiable.
\\

\multicolumn{4}{l}{\textbf{Causal interventions and local trace}}\\[2pt]

Qwen2.5 intervention
&
Opposite signs of the frozen learned direction; each directional gap is tested
against norm-matched random spectral directions
&
Budgets $.025/.05/.10$:
future gaps $-.00536$, $-.00833$, $-.01726$;
empirical $p=.0615,.0308,.0154$.
Source-layer sink change $=0$.
&
Establishes a dose-consistent suppressive causal effect in Qwen2.5.
\\

GLM-Z1 intervention
&
Opposite signs of the learned direction vs.\ 64 matched random directions
&
Budgets $.025/.05/.10$:
natural gaps
$-.01525$, $-.02976$, $-.05022$;
random 95\% ranges
$[-.01205,.00946]$,
$[-.02773,.02174]$,
$[-.04574,.03433]$;
$p=.0308,.0308,.0462$.
&
Independent cross-model causal corroboration with state-dependent magnitude.
\\

GPT-OSS frozen confirmation
&
Frozen L10 direction, $\rho=.025$, evaluated on 30 fresh prompts; endpoint is
mean sink strength at full-attention layers 11/13/15
&
Natural effect $+.05625$ $[.04983,.06267]$;
30 positive / 0 zero / 0 negative;
random median $-.00269$;
random empirical range $[-.05538,.05548]$;
one-sided $p=.0308$;
two-sided $p=.0462$;
mean KL $.0107$;
maximum source-layer sink change $=0$.
&
Strongest frozen fresh-cohort causal confirmation.
\\

Qwen2.5 next-layer trace
&
Frozen layer-1 edit; compare opposite signs at layer~2
&
Position-0 attention difference
$-.02142$ $[-.02419,-.01846]$;
sink-score difference
$-.02714$ $[-.04143,-.01357]$;
$r(\Delta A,\Delta M_{qk})=.544$ $[.395,.671]$;
97.7\% sign agreement;
value-norm sign agreement 33.9\%.
&
Links the immediate suppressive effect to reduced next-layer position-0 QK
preference.
\\

Direction robustness
&
Re-estimate directions from reduced preparation data
&
Qwen2.5:
half-vs-full cosine median $.9947$
(empirical 2.5--97.5\% $[.9750,.9987]$);
disjoint-half cosine median $.9789$
$[.9022,.9947]$;
held-out near-minus-far projection positive in 2000/2000 splits.
GPT-OSS:
half-vs-full cosine median $.9858$
(range $[.9740,.9916]$);
half-data effect median $+.05634$ vs.\ full $+.05625$;
effect-retention median $1.002$ (range $[.987,1.006]$);
12/12 preserve sign.
&
Shows that the learned directions and the GPT-OSS effect are not fragile to
substantial data reduction.
\\

\multicolumn{4}{l}{\textbf{Cross-layer propagation and boundaries}}\\[2pt]

GPT-OSS propagation
&
Layer-8/9 upstream edits compared with the direct layer-10 edit at the same
downstream state
&
Upstream relative hidden norms about $3.5\%$ and $3.0\%$ vs.\ $2.5\%$ direct;
targeted spectral projection retains about 96\% and 100\%;
hidden-displacement cosine with the direct edit about $.65$ and $.81$.
&
Shows that norm and the targeted spectral component can survive while the full
hidden perturbation rotates and the causal effect weakens.
\\

Qwen2.5 propagation
&
Layer-0 upstream edit compared with the direct layer-1 edit
&
Direct relative norm $2.51\%$;
upstream displacement at layer~1 $2.84\%$;
about 24\% of the targeted spectral projection remains;
hidden-displacement cosine about $.18$.
&
Shows a second form of reshaping in which the targeted spectral component is
also strongly filtered.
\\

Long-context identifiability
&
Frozen $T=256$ timing audits
&
Qwen3:
499/500 first cross at layer~7.
Qwen2.5:
429/500 cross and all 429 first cross at layer~4; 71 do not cross.
GLM-Z1 retains distributed onset support and remains estimable.
&
Separates failure of the comparison from negative evidence about the spectral
phenomenon.
\\

Qwen3.6 recurrent boundary
&
Full-attention checkpoints under three context/query settings
&
$T=64,q_{\min}=16$: 45/50 cross.
$T=256,q_{\min}=16$: 27/50 cross, with 5/27 persisting to the next checkpoint.
$T=256,q_{\min}=64$: 17/45 cross, with 5/17 persisting.
Fixed-target $\Delta$AP estimates are not robust across query windows.
&
Shows that a single persistent sink-onset layer is not an architecture-invariant
target.
\\

\end{longtable}
\endgroup

Table~\ref{tab:claim-map} summarizes how these quantitative results map to the
paper's claims and where each conclusion stops.

\begin{table}[htbp]
\centering
\caption{Summary of the main claims, the experiments that support them, and the limits of those conclusions.}
\label{tab:claim-map}
\scriptsize
\setlength{\tabcolsep}{3.5pt}
\begin{tabular}{p{0.20\linewidth}p{0.45\linewidth}p{0.27\linewidth}}
\toprule
Claim & Main evidence & Limitation \\
\midrule
The hidden-state spectrum changes before a sink forms
& Same-layer $\Afeat$ versus $\Afeat+\Sfeat$ prediction; removal of threshold-derived attention features; prompt shuffling; nonlinear attention baselines; expanded Qwen3 cohort
& The primary estimate is pooled, model-level support is uneven, and the stricter Qwen3 post-audit holdout shows weaker cohort-dependent prediction. \\

The clearest prompt-specific signal appears about two to three layers before sink formation
& Full horizon grid; removal of model--layer means; evaluation inside matched groups; sustained-sink definitions; $T=256$ GLM-Z1 result
& The pooled signal extends farther, and two to three layers is not a universal causal window. \\

The spectral change is broader than one measured extreme activation
& Compact outlier controls; capacity-matched activation measurements; random projections; Qwen3 timing experiment
& These prediction tests do not identify a unique causal mechanism. \\

Changing the spectrum can change later sink formation
& Post-layer singular-value edits; norm-matched random directions; Qwen2.5 and GLM-Z1 results; frozen 30-prompt GPT-OSS confirmation; half-data direction stability
& The sign and size depend on model and state, and larger edits can cause broader functional changes. \\

The immediate Qwen2.5 suppression enters the next attention computation through a lower position-0 QK preference
& Layer-1 post-block edit; layer-2 prompt--head QK-margin trace; source-attention invariance
& This is a local diagnostic in one model and does not establish a universal QK mechanism. \\

Later transformer layers change the perturbation
& GPT-OSS layer-8/9/10 propagation; Qwen2.5 layer-0-to-1 propagation
& The cross-layer experiment covers two model families and is not claimed as a universal law. \\

The same experiment cannot be applied in every setting
& Qwen3 and Qwen2.5 long-context timing collapse; GLM-Z1 long-context comparison; fresh Qwen3/GLM-4 support tests; GPT-OSS and Qwen3.6 architecture-specific cases
& Lack of a usable comparison does not show that the underlying spectral phenomenon is absent. \\
\bottomrule
\end{tabular}
\end{table}

Taken together, these checks support the same conclusion as the main text:
\textbf{attention sinks are preceded by a short spectral reorganization of
hidden states, and directly perturbing this geometry can causally alter
subsequent sink formation.}

The additional analyses also define the limits of this statement. The
approximately two--three-layer timing is an observational result rather than a
universal causal window; the size of the predictive gain is cohort dependent;
the spectrum is not a complete description of the hidden state; and some
architectures or prompt distributions do not provide enough variation to define
the same sink-formation problem.

\end{document}